\pdfoutput=1
\PassOptionsToPackage{table}{xcolor}

\documentclass[11pt]{article}
\usepackage{microtype}
\usepackage{graphicx}
\usepackage{subcaption}
\usepackage{booktabs} 

\usepackage[final]{acl}
\usepackage{times}
\usepackage{latexsym}
\usepackage{inconsolata}

\usepackage[T1]{fontenc}
\usepackage[utf8]{inputenc}

\usepackage{amsmath}
\usepackage{amssymb}
\usepackage{mathtools}
\usepackage{amsthm}

\usepackage[utf8]{inputenc} 
\usepackage[T1]{fontenc}    
\usepackage{url}            
\usepackage{booktabs}       
\usepackage{amsfonts}       
\usepackage{nicefrac}       
\usepackage{microtype}      
\usepackage{xspace}
\usepackage{soul}
\usepackage{caption}
\usepackage{siunitx}
\usepackage{bbm}

\usepackage{tabularx}
\definecolor{mydarkblue}{RGB}{0, 0, 139}
\hypersetup{
    colorlinks=true,
    linkcolor=mydarkblue,
    filecolor=mydarkblue,
    citecolor=mydarkblue,
    urlcolor=mydarkblue,
}
\usepackage{gensymb}

\usepackage{mathtools}
\usepackage{bm}
\usepackage[capitalize, noabbrev, nameinlink]{cleveref}
\usepackage{graphicx}
\usepackage{algorithm}
\usepackage{enumitem}
\usepackage{wrapfig}
\usepackage{pifont}
\usepackage{multirow}
\usepackage{tabularx}
\usepackage{placeins}

\usepackage{amsmath,amsfonts,bm}

\def\eqref#1{equation~\ref{#1}}

\def\1{\bm{1}}

\DeclareMathAlphabet{\mathsfit}{\encodingdefault}{\sfdefault}{m}{sl}
\SetMathAlphabet{\mathsfit}{bold}{\encodingdefault}{\sfdefault}{bx}{n}

\usepackage{hyperref}
\usepackage{url}
\usepackage{mathtools}
\usepackage{wasysym}
\usepackage{booktabs}
\usepackage{tcolorbox}
\usepackage{listings}

\newcommand{\Algname}{$\textsc{MSR}$\xspace}
\newcommand{\molgiven}{analytic reasoning\xspace}
\newcommand{\molinfer}{synthetic reasoning\xspace}

\definecolor{byzantium}{rgb}{0.44, 0.16, 0.39}
\definecolor{grn}{rgb}{0.1, 0.6, 0.1}
\definecolor{mgt}{rgb}{0.7, 0.3, 0.7}
\definecolor{chamoisee}{rgb}{0.63, 0.47, 0.35}
\definecolor{purp}{rgb}{0.65, 0.16, 0.65}
\definecolor{alizarin}{rgb}{0.82, 0.1, 0.26}
\definecolor{azure(colorwheel)}{rgb}{0.0, 0.5, 1.0}
\definecolor{brown}{rgb}{0.65, 0.16, 0.16}
\definecolor{whitegray}{RGB}{243, 243, 243}

\definecolor{myblue}{RGB}{67, 132, 194}
\definecolor{myred}{RGB}{239, 76, 86}
\definecolor{myyellow}{RGB}{248, 193, 89}
\definecolor{mygreen}{RGB}{65, 188, 158}
\definecolor{MyLightGray}{RGB}{200,200,229}
\definecolor{MyGray}{RGB}{100, 100, 100}

\usepackage[capitalize,noabbrev]{cleveref}

\title{Structural Reasoning Improves Molecular Understanding of LLM}

\author{Yunhui Jang \\ KAIST \\
  \texttt{yunhuijang@kaist.ac.kr} \\\And
  Jaehyung Kim \\
  Yonsei University\\
  \texttt{jaehyungk@yonsei.ac.kr} \\ \\\And
  Sungsoo Ahn \\
    KAIST \\
    \texttt{sungsoo.ahn@kaist.ac.kr}}

\begin{document}
\maketitle

\begin{abstract}
Recently, large language models (LLMs) have shown significant progress, approaching human perception levels. In this work, we demonstrate that despite these advances, LLMs still struggle to reason using molecular structural information. This gap is critical because many molecular properties, including functional groups, depend heavily on such structural details. To address this limitation, we propose an approach that sketches molecular structures for reasoning. Specifically, we introduce \textbf{M}olecular \textbf{S}tructural \textbf{R}easoning (\textbf{MSR}) framework to enhance the understanding of LLMs by explicitly incorporating the key structural features. We present two frameworks for scenarios where the target molecule is known or unknown. We verify that our \Algname improves molecular understanding through extensive experiments.
\end{abstract}

\section{Introduction}

Large language models~\citep[LLMs;][]{touvron2023llamaopenefficientfoundation, openai2024gpt4technicalreport, raffel2020t5} have demonstrated remarkable performance across various tasks. To leverage their potential in chemistry, several prior works~\citep{edwards2022molt5, christofidellis2023chemt5, fang2024molinstructions, pei-etal-2023-biot5} have proposed chemical LLMs (i.e., specialized LLMs pre-trained on both natural language and molecular representations) for molecular tasks such as molecule captioning, description-based molecule generation~\citep{edwards2022molt5}, and retrosynthesis \citep{fang2024molinstructions}.

However, chemical LLMs still struggle to fully understand the molecular structure \citep{ganeeva2024lost,white2023assessment}. This is critical since structure-based reasoning plays an important role in many molecular tasks. For instance, chemists often consider a molecule toxic if it contains a phenol group, as phenoxyl radicals can form and interact with biological membranes \citep{hansch2000phenol}. 
\textcolor{black}{This becomes even more evident in real-world applications of LLMs. As demonstrated in \cref{fig: failure_case}, injecting accurate structural information into the model can slightly improve its ability to generate correct molecules. This highlights the importance of explicitly incorporating structural reasoning into LLMs.}

{
\textcolor{black}{To address this aspect, we consider a framework for LLMs to first reason about the molecular structure for molecular tasks, similar to how LLMs improve arithmetic and commonsense tasks through intermediate reasoning steps~\citep{wei2023chainofthoughtpromptingelicitsreasoning, kojima2022large}. A na\"ive approach is to prompt LLMs to infer the structural information before generating an answer. However, we find this to be ineffective since even the state-of-the-art LLMs~\citep{openai2024gpt4technicalreport, touvron2023llamaopenefficientfoundation} struggle to accurately capture essential molecular structures, as described in \cref{fig: analysis} and \cref{subsec: failure}. 
}

\begin{figure*}[t]
    \centering
    \hspace{1em}%
    \begin{subfigure}[t]{0.6\textwidth}
        \includegraphics[width=\linewidth]{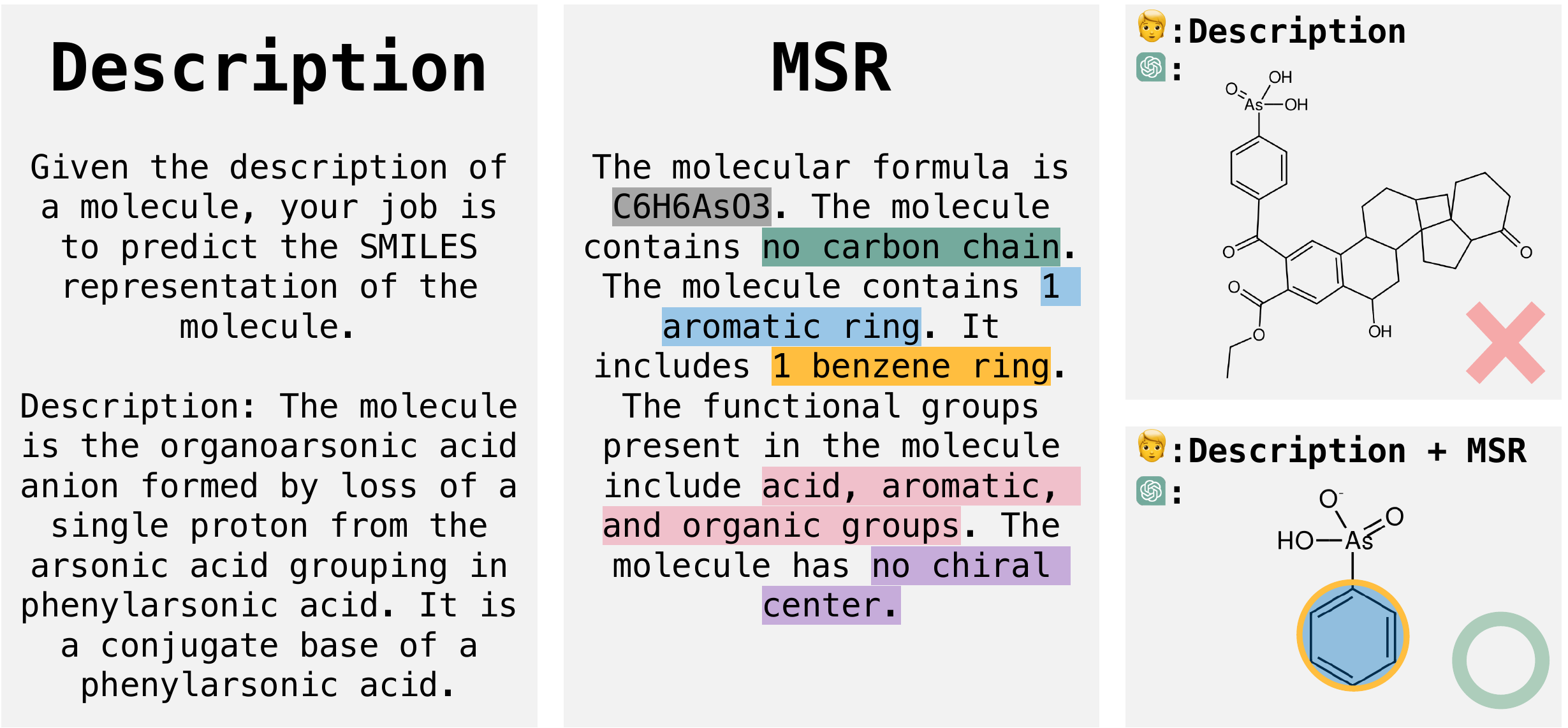}
        \vspace{-0.1in}
        \captionsetup{justification=centering}
        \caption{Incorporating \Algname improves GPT-4o in molecule generation.}\label{fig: failure_case}
    \end{subfigure}
    \centering
    \begin{subfigure}[t]{0.36\textwidth}
        \includegraphics[width=\linewidth]{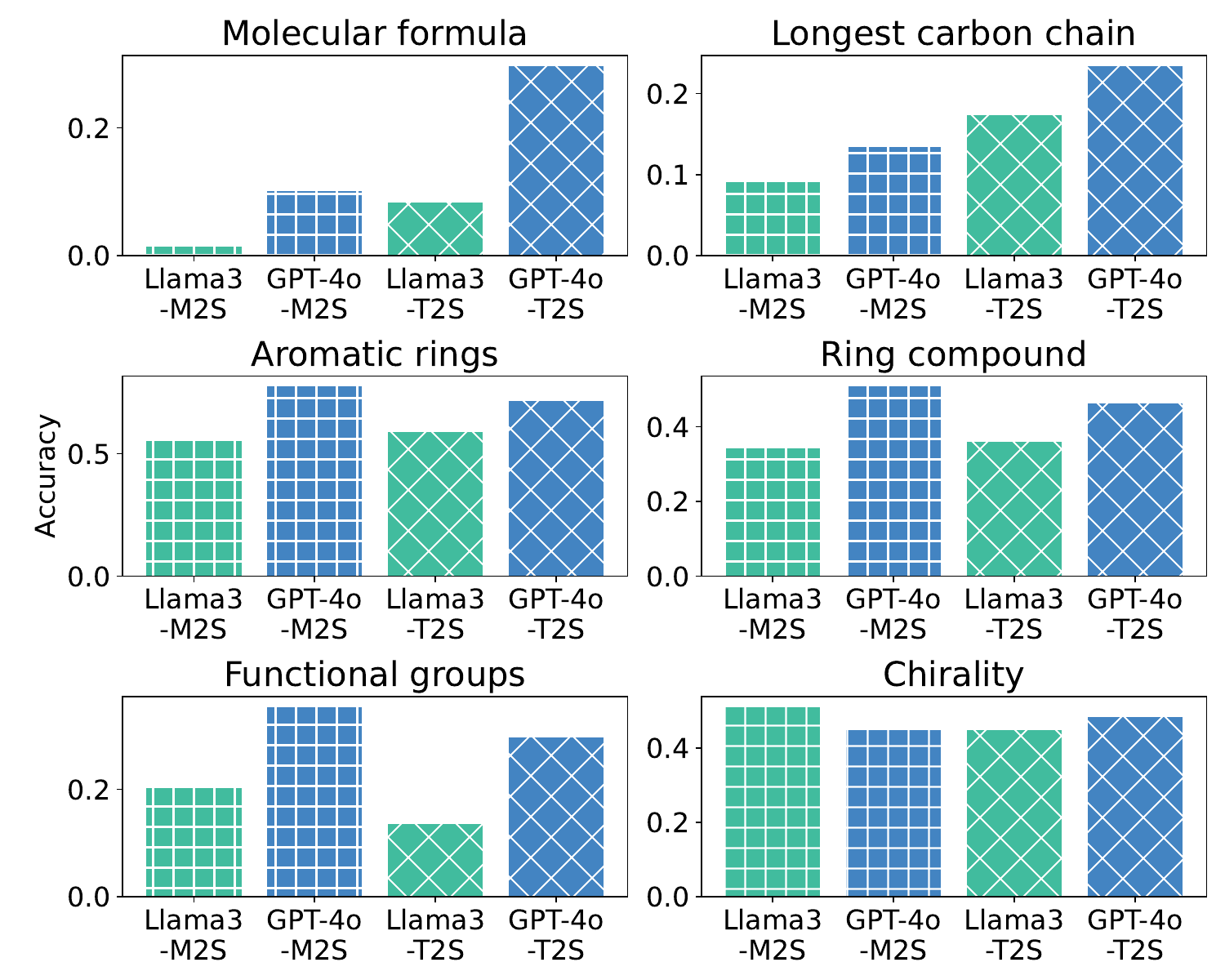}
        \vspace{-0.1in}
        \captionsetup{justification=centering}
        \caption{LLMs' capability of structural inference.}\label{fig: analysis}
    \end{subfigure}
    
    \caption{\textbf{Overview of LLMs with structural information}. (\subref{fig: failure_case}) Each color in \Algname represents a structural component. The top molecule is incorrectly generated using only the description while the bottom is correctly generated by incorporating the description and \Algname. (\subref{fig: analysis}) Despite the importance of structural information, even recent LLMs struggle to accurately infer key structural details from molecular representations such as SMILES (Molecule-to-structure; M2S) or given descriptions (Text-to-structure; T2S).}\label{fig: llm}
    \vspace{-0.2in}
\end{figure*}

{
In this paper, we propose \Algname, a simple yet general framework for \textbf{M}olecular \textbf{S}tructural \textbf{R}easoning that progressively sketches the structural features of molecules to improve molecular task performance. To this end, we identify key structural elements crucial for the reasoning of LLMs to solve molecular tasks. Moreover, we propose fine-tuning procedures that employ external tools to identify the molecular structural information.
}

\textcolor{black}{
In particular, our frameworks to fine-tune LLMs for molecular structural reasoning are designed for both molecular and non-molecular inputs. A framework consists of \textit{reasoning module} and an \textit{answering module}. The reasoning module generates structural information to enhance the understanding of the molecule. The answering module generates the final answer based on the original input and the output of the reasoning module. The overall frameworks are illustrated in \cref{fig: training}. 
}

\textcolor{black}{To be specific, we consider two types of reasoning framework, inspired by how humans generally form knowledge via analysis and synthesis~\citep{critique1899kant}. On the one hand, \textit{analysis} refers to breaking down complex information into fundamental components for better understanding. In molecular tasks, \molgiven applies when the molecule is provided as input, allowing the model to decompose its structure for meaningful insights. Specifically, we utilize external tools like RDKit~\citep{greg2024rdkit} as the reasoning module to precisely extract structural information from the molecule.}

\textcolor{black}{On the other hand, \textit{synthesis} constructs a whole from its constituent parts. This aligns with molecular tasks where the molecule must be generated from non-molecular input, e.g., textual description, requiring the model to infer structural information and reconstruct the entire molecule. In detail, for \molinfer, we fine-tune the LLMs as the reasoning module that generates \Algname \citep{ho-etal-2023-large, fu2023specializing, magister-etal-2023-teaching} based on the given input. We additionally incorporate a novel \textit{matching-ratio-based rejection sampling} into the answering module, to ensure that the generated molecule aligns with \Algname, using the external tools for validation.}

We empirically show that incorporating \Algname into chemical LLMs \citep{edwards2022molt5, christofidellis2023chemt5} and general LLMs~\citep{touvron2023llamaopenefficientfoundation, openai2024gpt4technicalreport} both lead to consistent performance improvements in three molecular tasks: molecule-to-text, retrosynthesis, and text-to-molecule. In particular, chemical LLMs outperform the considered baselines when combined with our MSR framework. In summary, our contributions are as follows:

\begin{itemize}[leftmargin=.2in]
    \item We identify and evaluate the limits of LLMs in inferring molecular structural information.
    \item We propose \Algname, a simple yet broadly applicable molecular reasoning framework that progressively sketches molecular structures.
    \item We introduce an \molgiven for \Algname when the input molecule is given, leveraging external tools for structural identification.
    \item We develop a \molinfer for \Algname when the molecule is in the desired output, incorporating fine-tuning for the reasoning module and a novel matching ratio-based rejection sampling procedure for the answering module.
    \item We validate the effectiveness of \Algname by demonstrating consistent performance improvements across chemical and general LLMs.
\end{itemize}

\section{Recent large language models do not understand structural information}\label{subsec: failure}

Here, we demonstrate that even the recent LLMs, i.e., GPT-4o~\citep{openai2024gpt4technicalreport} and Llama3-8B-Instruct~\citep{touvron2023llamaopenefficientfoundation}, fail to infer important structural information from the given inputs, such as molecular representations (e.g., SMILES~\citep{weininger1988smiles}) and the text descriptions~\citep{edwards-etal-2021-text2mol}. Notably, such tasks can be considered straightforward for individuals with a bachelor's degree in chemistry. 

\begin{figure}[t]
    \centering
    \includegraphics[width=\linewidth]{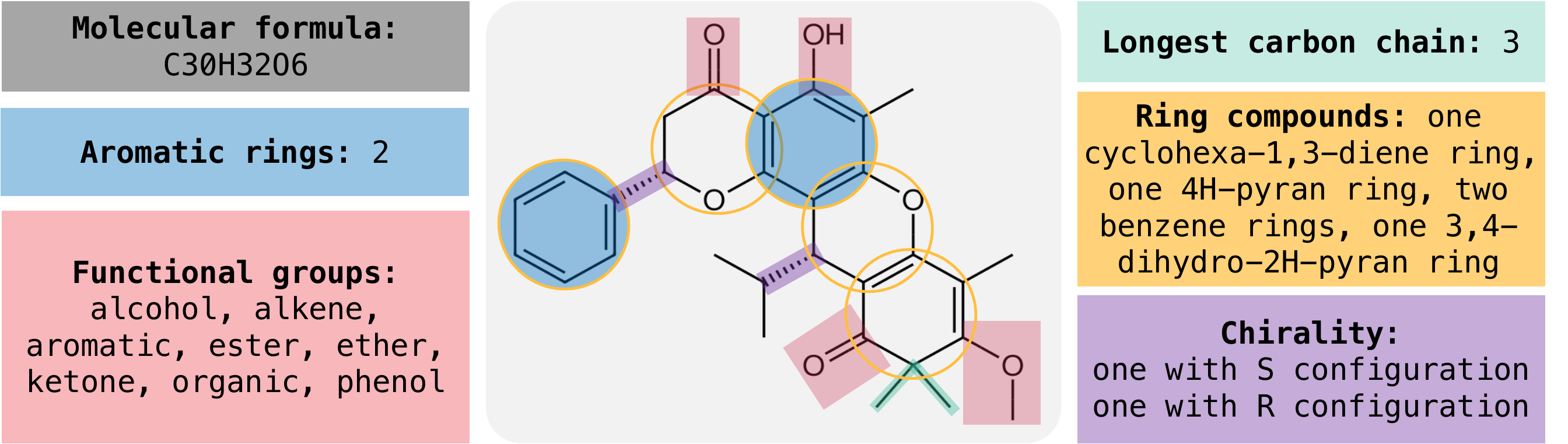}
    \caption{\textbf{The six key structural information: \textcolor{gray}{molecular formula}, \textcolor{mygreen}{longest carbon chain length}, \textcolor{myblue}{aromatic rings}, \textcolor{myyellow}{ring compounds}, \textcolor{myred}{functional groups}, and \textcolor{violet}{chirality}.} The same color indicates the structural information and the corresponding component of the molecule.
    }\label{fig: cot_structure}
    \vspace{-0.2in}
\end{figure}

Our analysis is inspired by how chemists reason about the structure to analyze a molecule. They progressively identify the molecular structure, starting with primary elements like rings and long carbon chains before identifying finer details such as functional groups. Reflecting on this behavior, we define six significant key structural elements for chemical reasoning as illustrated in \cref{fig: cot_structure}. In detail, these six key structural components include (1) molecular formula, (2) longest carbon chain length, (3) aromatic rings, (4) ring compounds, (5) functional groups, and (6) chirality.

\paragraph{Molecular formula.} The molecular formula provides essential information about a molecule’s composition, specifying the number and type of atoms present. This information is critical because, for example, it directly determines the molecular weight. To illustrate, although 2-Butanol (C$_4$H$_{10}$O) and 2-Propanol (C$_3$H$_8$O) are composed of the same type of atoms, i.e., carbon, hydrogen, and oxygen, their differing molecular formulas result in distinct molecular weights (74.1g/mol for 2-Butanol and 60.1g/mol for 2-Propanol). These differences lead to the change in boiling points, $99.4^{\circ}$C and $82.3^{\circ}$C, respectively, as shown in the gray part of \cref{fig: structure}.

\paragraph{Longest carbon chain.} The longest carbon chain (excluding atoms in ring systems) forms the molecular backbone where functional groups are attached. The length of this chain significantly influences properties like solubility. For example, extending the carbon chain of 2-Butanol from four to six carbons creates 2-Hexanol, which exhibits reduced solubility. This is illustrated in the green section of \cref{fig: structure}.

\paragraph{Aromatic rings.} Aromatic rings (e.g., benzene and pyridine) play a critical role in determining the stability and electronic properties. For instance, adding a benzene ring to 2-Butanol yields 1-Phenyl-2-Propanol, which has enhanced stability and greater oxidation resistance. This transformation is shown in the blue section of \cref{fig: structure}.

\paragraph{Ring compounds.} Similar to the longest carbon chain, ring structures often serve as the backbone where functional groups are attached. The ring system significantly affects molecular behavior and reactions. For example, although 2-Butanol and Cyclobutanol share the same number of carbons and oxygen, the ring in Cyclobutanol introduces a tendency toward ring-opening reactions, as depicted in the yellow section of \cref{fig: structure}.

\begin{figure}[t]
\vspace{-0.2in}
    \centering
    \includegraphics[width=\linewidth]{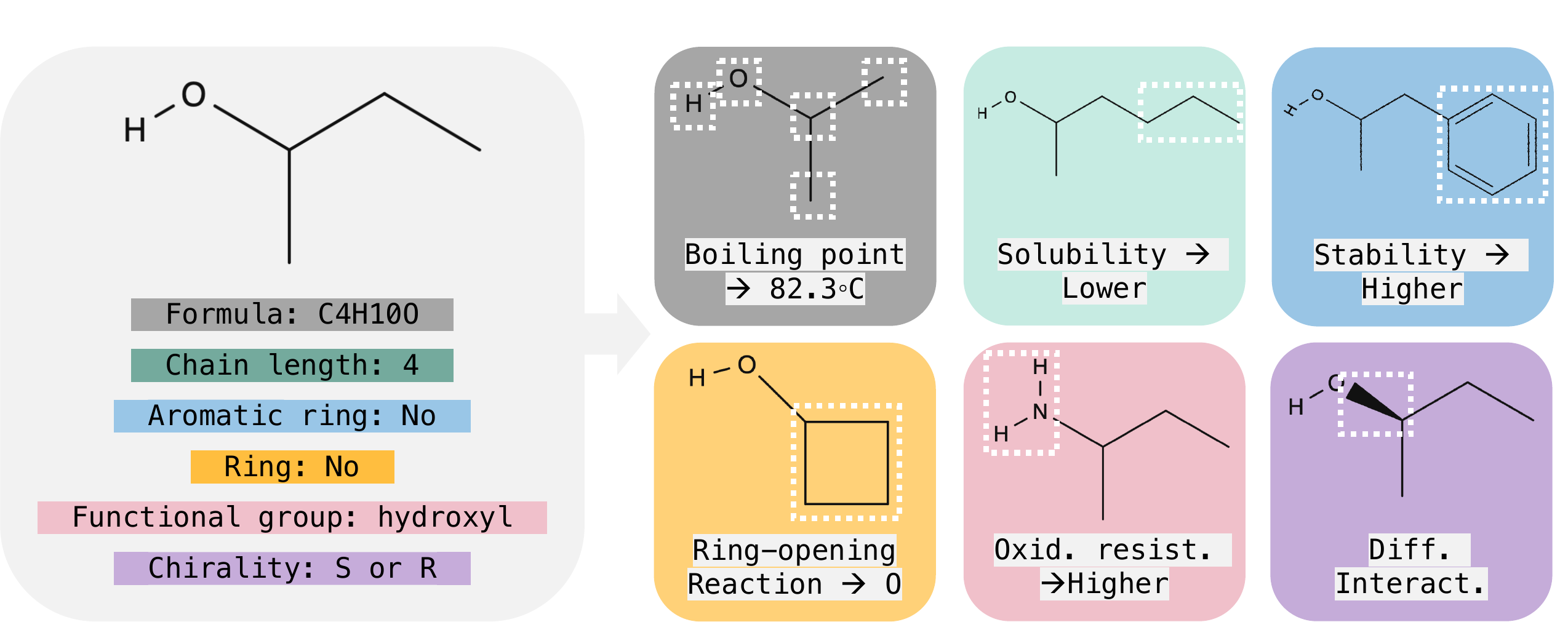}
    \caption{\textbf{Illustration of the importance of structural information.} This example shows how replacing each structural information (dashed box) alters the molecule. Colors correspond to the structural elements in \cref{fig: cot_structure}.
    }\label{fig: structure}
    \vspace{-0.2in}
\end{figure}

\paragraph{Functional groups.} Functional groups, e.g., hydroxyl, amino, ester, etc., play a pivotal role in determining the chemical reactivity. For example, alcohols with a hydroxyl group (-OH) are prone to oxidize more while the molecules with an amino group (-NH$_2$) are generally resistant to oxidation under mild conditions. A single replacement of a hydroxyl (-OH) group in 2-Butanol with an amino (-NH$_2$) group leads to 2-Butanamine, which has increased oxidation resistance, as described in the red part of \cref{fig: structure}.

\paragraph{Chiral centers.} Chirality refers to the stereochemical property of a molecule that makes it non-superimposable on its mirror image, leading to different chemical behaviors. The chirality is determined by the chiral centers and their configurations, i.e., R- and S-configuration \footnote{The names of R and S come from the Latin word \textit{Rectus} and \textit{Sinister}, which means right and left, respectively.}, which describe the spatial arrangement of the groups around the chiral centers. This leads to different interactions between other molecules with chirality. For instance, (R)-2-Butanol and (S)-2-Butanol may interact differently with other chiral substances. This is described in the purple part of \cref{fig: structure}.

\textcolor{black}{Despite their significance, we observe that even recent LLMs often fail to accurately infer crucial structural details from the molecule or their text description. Specifically, as shown in \cref{fig: analysis}, both GPT-4o and LlaMA3-8B-Instruct fail to capture the structural information accurately when the molecule is provided (Molecule-to-structure; M2S) or the description is provided (Text-to-structure; T2S). Notably, we provide detailed experimental settings and prompts in \cref{appx: exp_anal}.}

\textcolor{black}{First, when provided with a molecule (M2S), both GPT-4o and LlaMA3-8B-Instruct struggled to achieve high accuracy. Even in their best-performing case, counting the number of aromatic rings, their accuracies remained low, at approximately 50\% and 75\%, respectively. Similarly, when given a detailed text description (T2S), both models still failed to achieve high accuracy. This implies that LLMs struggle to fully understand the molecular structures, whether presented as molecular representations or text descriptions. These observations highlight the potential benefits of explicitly incorporating structural reasoning to enhance molecular comprehension.}

\section{\Algname: Molecular Structural Reasoning}\label{sec: method_cot}

\begin{figure*}[t]
    \centering
    \hspace{1em}%
    \begin{subfigure}[t]{0.48\textwidth}
        \includegraphics[width=\linewidth]{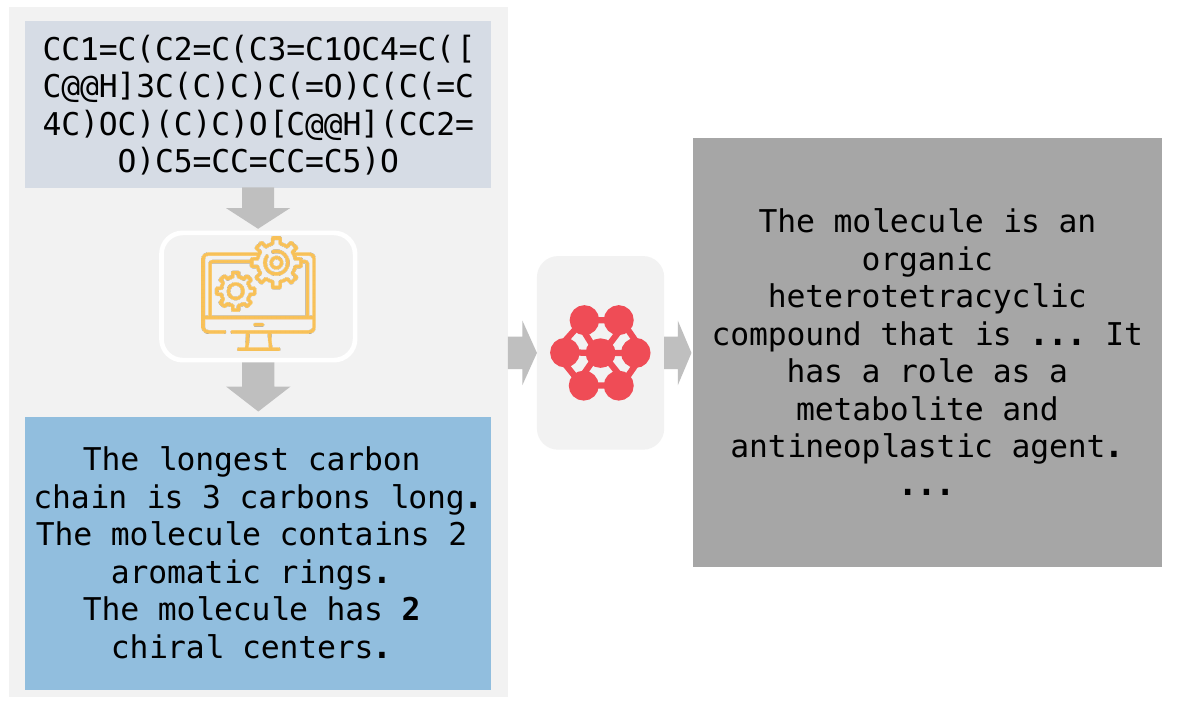}
        \vspace{-0.1in}
        \captionsetup{justification=centering}
        \caption{Analytic reasoning}\label{fig: train_given_mol}
    \end{subfigure}
    \centering
    \begin{subfigure}[t]{0.48\textwidth}
        \includegraphics[width=\linewidth]{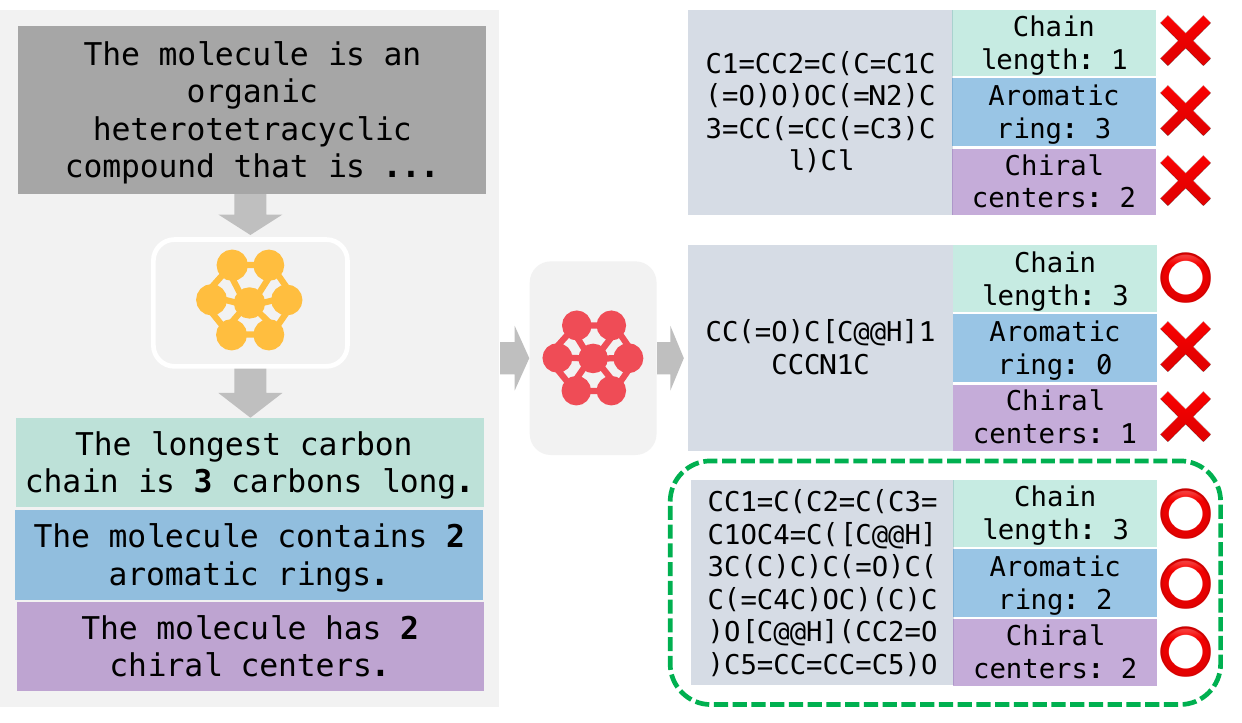}
        \vspace{-0.1in}
        \captionsetup{justification=centering}
        \caption{Synthetic reasoning}\label{fig: train_wo_mol}
    \end{subfigure}
    
    \caption{\textbf{Overview of \Algname fine-tuning framework.} Analytic reasoning applies when the input molecule is available, while synthetic reasoning applies when it is not. \textcolor{MyLightGray}{Light gray} boxes denote the molecules (SMILES); \textcolor{MyGray}{gray} boxes denote related description; colored boxes represent \Algname. The \textcolor{myyellow}{yellow} and the \textcolor{myred}{red} modules perform reasoning and answering, respectively. In (\subref{fig: train_given_mol}), \textcolor{myyellow}{yellow} module indicates an external tool.
    In (\subref{fig: train_wo_mol}), colors indicate \Algname and the corresponding structural elements; here, the third molecule is chosen after \textit{matching ratio-based rejection sampling} according to its highest matching ratio (3/3).}\label{fig: training}
    \vspace{-0.1in}
\end{figure*}

\textcolor{black}{Here, we present our framework for enhancing LLMs' understanding of molecules through \textbf{M}olecular \textbf{S}tructural \textbf{R}easoning (\Algname). \Algname incorporates six key structural elements as reasoning for LLMs, following a two-stage process~\citep{zhang2024multimodal}: a reasoning stage and an answering stage. First, a \textit{reasoning module} generates \Algname, providing supplementary structural information for a better understanding of the molecule. Next, an \textit{answering module} generates the final output using the input augmented with the generated \Algname. The framework is illustrated in \cref{fig: training}.}

\textcolor{black}{To address various tasks, we categorize the \Algname framework based on whether the molecule is provided as input (\textit{\molgiven}) or must be inferred as output (\textit{\molinfer}). In summary, for \molgiven, one decomposes complex molecules into fundamental structural components to better understand their structure. For \molinfer, one constructs the entire molecule from its constituent structural components.}

\definecolor{whitegray}{RGB}{243, 243, 243} 
\begin{table}[t]
  \centering
  \resizebox{\linewidth}{!}{
\begin{tabular}{ccc}
    \toprule[1.25pt]
    \textbf{Component} & \textbf{Expression} & \textbf{Description}  \\
    \midrule[1.25pt]
  \textcolor{gray}{\textbf{Molecular}} & \textit{The molecular formula is} & $X_m$: $m$-th atom type \\
  \textcolor{gray}{\textbf{formula}} & \textit{ $X_{1}N_{1} \cdots X_{M}N_{M}$.} & $N_m$: \# of $m$-th atoms \\
  \midrule
 \textcolor{mygreen}{\textbf{Longest carbon}} & \textit{The longest carbon chain} & $N$: the length of  \\
 \textcolor{mygreen}{\textbf{chain length}} & \textit{is $N$ carbons long.} & the longest carbon chain  \\
 \midrule
\textcolor{myblue}{\textbf{Aromatic}} & \textit{The molecule contains} & $N$: \# of \\
 \textcolor{myblue}{\textbf{rings}} & \textit{$N$ aromatic rings.} & aromatic rings \\
  \midrule
 \textcolor{myyellow}{\textbf{Ring}} & \textit{It includes $N_1$ $X_1$ rings,} & $N_m$: \# of $m$-th ring \\
 \textcolor{myyellow}{\textbf{compounds}} & \textit{$\cdots$, $N_{M}X_{M}$ ring(s).} & $X_m$: IUPAC name of $m$-th ring \\
 \midrule
 \textcolor{myred}{\textbf{Functional}} & \textit{The functional groups include} & \multirow{2}{*}{$X_m$: the name of functional group} \\
\textcolor{myred}{\textbf{groups}} & \textit{ $X_1$, $\cdots$, and $X_M$ group.} & \\
 \midrule
 \multirow{3}{*}{\textcolor{violet}{\textbf{Chirality}}} & \textit{The molecule has $N$ chiral} & $N_S$: \# of chiral centers of $S$ config. \\
 & \textit{ centers: $N_S$ with $S$ configuration} & $N_R$: \# of chiral centers of $R$ config. \\
 & \textit{and $N_R$ with $R$ configuration.}& $N=N_S+N_R$ \\
    \bottomrule[1.25pt]
    \end{tabular}
    }
    \vspace{-0.1in}
    \caption{\textbf{The description of each component of MSR.}  \label{tab:reason_comp}}
    \vspace{-0.2in}
\end{table}


\subsection{Overview of \Algname}\label{subsec: method_cot}

Here, we introduce \Algname, a molecular structural reasoning framework that enhances language models’ understanding of molecules. Each component of \Algname corresponds to one of the six structural elements introduced in \cref{subsec: failure}. \textcolor{black}{The expression and corresponding description of the reasoning for each structural component in \Algname are provided in \cref{tab:reason_comp}.} Additionally, a concrete example illustrating \Algname is shown in \cref{fig: cot_structure}.

\subsection{Analytic reasoning}\label{subsec: method_m2t}

In MSR, analytic reasoning refers to decomposing a given input molecule into smaller structural components for enhanced comprehension. When the input molecule is available, one can utilize a deterministic reasoning module for its decomposition. Our approach integrates \Algname by (1) employing external tools like RDKit~\citep{greg2024rdkit} to extract precise structural information as a reasoning module, and (2) fine-tuning the answering module LLM with the generated rationale as an additional input. The overall workflow is described in \cref{fig: train_given_mol}.

\paragraph{Reasoning module.} In the \molgiven scenario, we employ external tools to extract precise structural information from the input molecule. This process eliminates uncertainty, as the structural information is deterministic for a given molecule. Next, this information serves as \Algname, which guides the answering module.

\paragraph{Answering module.}
With the molecule and its corresponding \Algname as input, we fine-tune the chemical LLMs to generate the desired output of the molecule. In our experiments, we mainly consider MolT5~\citep{edwards2022molt5} and ChemT5~\citep{christofidellis2023chemt5}, as the answering module.

\subsection{Synthetic reasoning}\label{subsec: method_t2m}
Synthetic reasoning refers to composing structural information to construct an entire molecule. When the input molecule is unavailable, the relevant structural information must first be inferred before generating the final molecule. To address this, we fine-tune a reasoning module to generate \Algname, which is then attached to the input and utilized by the answering module to generate the final molecule, as illustrated in \cref{fig: train_wo_mol}. 

\paragraph{Reasoning module.} We fine-tune the chemical LLMs to generate \Algname \textcolor{black}{similar to prior works that fine-tune LLMs to generate chain-of-thoughts~\citep{ho-etal-2023-large, fu2023specializing, magister-etal-2023-teaching}.} Unlike \molgiven, where external tools can precisely extract structural information, the reasoning module in \molinfer must infer this information from the input.

\definecolor{whitegray}{RGB}{243, 243, 243} 

\begin{table}[t]
  
  \centering
  \resizebox{\linewidth}{!}{
\begin{tabular}{cccccccc}
    \toprule[1.25pt]
      & {\textbf{BL.2} } & {\textbf{BL.4} } &{\textbf{RO.1} } &{\textbf{RO.2} } &{\textbf{RO.L} } &{\textbf{ME.} } \\
    \midrule[1.25pt]
    \rowcolor{whitegray} \multicolumn{7}{l}{\textit{Baselines (without reasoning)}} \\
    \midrule
    Meditron-7B & 0.792 & 0.576 & 0.797 & 0.602 & 0.575 & 0.757 \\
    \midrule
    Mol2Lang-VLM & 0.777 & 0.563 & 0.786 & 0.591 & 0.565 & 0.741 \\ 
    \midrule
    BioT5+ & 0.798 & 0.579 & 0.812 & 0.617 & 0.584 & 0.777 \\
    \midrule[1.25pt]
    \rowcolor{whitegray} \multicolumn{7}{l}{\textit{Chemical LLMs (fine-tuning)}} \\
    \midrule
    MolT5-small & 0.709 & 0.512 & 0.745 & 0.558 & 0.544 & 0.701 \\
    \quad+\Algname & \textcolor{teal}{0.780} & \textcolor{teal}{0.565} & \textcolor{teal}{0.807} & \textcolor{teal}{0.613} & \textcolor{teal}{0.585} & \textcolor{teal}{0.757}  \\
    \midrule
    MolT5-base & 0.738 & 0.535 & 0.750 & 0.559 & 0.539 & 0.718  \\
    \quad+\Algname & \textcolor{teal}{0.805} & \textcolor{teal}{0.592} & \textcolor{teal}{0.864} & \textcolor{teal}{0.677} & \textcolor{teal}{0.642} & \textcolor{teal}{0.822}  \\
    \midrule
    MolT5-large & 0.769 & 0.556 & 0.777 & 0.580 & 0.557 & 0.743 \\
    \quad +\Algname & \textcolor{teal}{\textbf{0.832}} & \textcolor{teal}{\textbf{0.622}} & \textcolor{teal}{\textbf{0.914}} & \textcolor{teal}{\textbf{0.743}} & \textcolor{teal}{\textbf{0.691}} & \textcolor{teal}{\textbf{0.878}} \\
    \bottomrule[1.25pt]
    \end{tabular}
    }
    \vspace{-0.1in}
    \caption{\textcolor{black}{\textbf{Molecule-to-text performance for L+M val.} BL., RO., and ME. stand for BLEU, ROUGE, and METEOR, respectively.}}\label{tab:mol2text_lm}
    \vspace{-0.2in}
\end{table}

Notably, we selectively retain only reliable structural components before incorporating them into the answering module. One considers the component to be reliable if it achieves sufficiently high reasoning accuracy across the entire dataset. This selection process also leverages the deterministic nature of structural information, allowing a quantitative evaluation of the reasoning module's capability in generating each type of rationale, as presented in \cref{subsec: mol_gen}.

\paragraph{Answering module.} Similar to the \molgiven scenario, we fine-tune chemical LLMs to generate an appropriate molecule from the input and its corresponding \Algname. To further enhance structural consistency between the generated molecule and \Algname, we propose a \textit{matching ratio-based rejection sampling} method.

\textcolor{black}{
Specifically, the model first generates $k$ candidate molecules using beam search. Then, for each candidate, one computes the matching ratio, which quantifies the consistency between the generated molecule's structural components and those in \Algname. Finally, the molecule with the highest matching ratio is selected as the final output, ensuring the consistency between the rationale and the generated answer.}

\textcolor{black}{Again, this approach leverages the deterministic nature of the molecular structural information, allowing us to easily obtain the information with external tools and compare them between the rationale and the generated molecule. Notably, the search process differs from prior works~\citep{wang2023selfconsistency, xi-etal-2023-self, sun2024itercot} that search over rationale-answer pairs since our method focuses on searching the answer that coincides with a given rationale.
}

\section{Experiments}\label{sec: exp}

\begin{figure*}[t]
    \centering
    \includegraphics[width=0.98\linewidth]{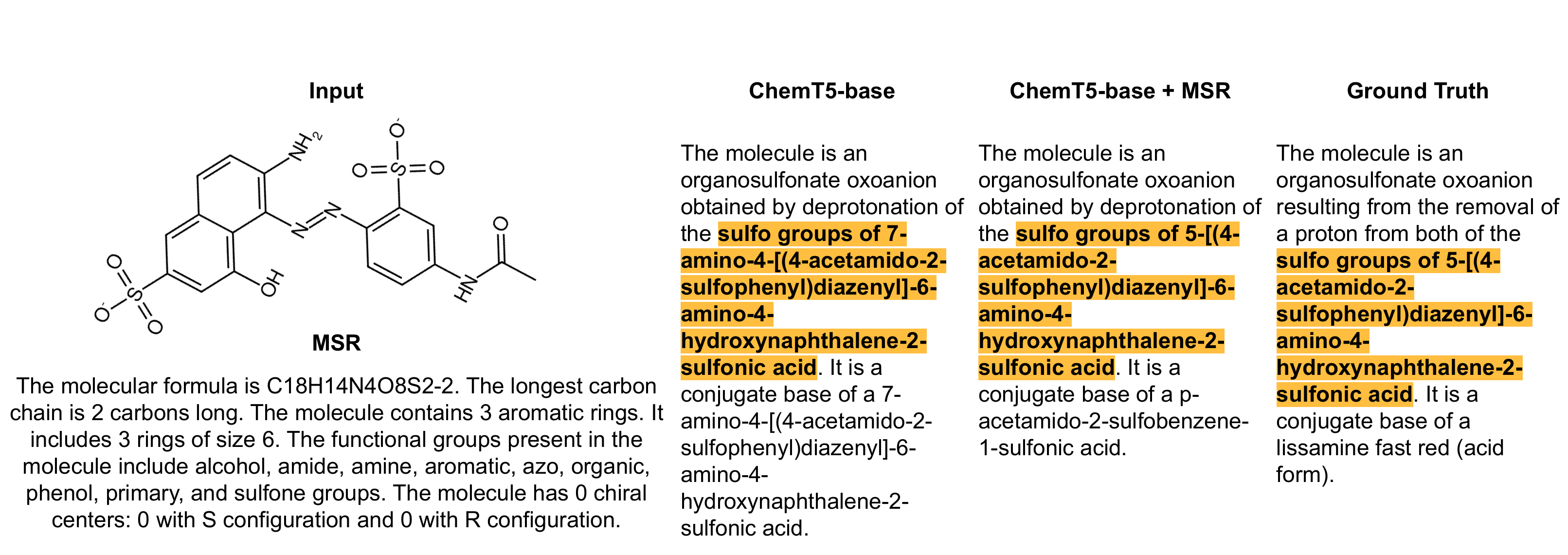}
    \vspace{-0.15in}
    \caption{\textbf{An example of generated samples for molecule-to-text.} We observe that \Algname improves the accuracy of detailed molecular information (highlighted in yellow). We provide more examples in \cref{appx: sample}.}\label{fig: example}
    \vspace{-0.2in}
\end{figure*}

\definecolor{whitegray}{RGB}{243, 243, 243} 

\begin{table}[t]
  \centering
  \resizebox{\linewidth}{!}{
\begin{tabular}{ccccccc}
    \toprule[1.25pt]
      & {\textbf{BL.2}} & {\textbf{BL.4}} &{\textbf{RO.1}} &{\textbf{RO.2}} &{\textbf{RO.L}} &{\textbf{ME.}} \\
    \midrule[1.25pt]
   \rowcolor{whitegray} \multicolumn{7}{l}{\textit{Baselines (without reasoning)}} \\
    \midrule
    T5-base & 0.511 & 0.423 & 0.607 & 0.451 & 0.550 & 0.539 \\
    \midrule
    MolXPT & 0.594 & 0.505 & 0.660 & 0.511 & 0.597 & 0.626 \\
    \midrule
    BioT5 & 0.635 & 0.556 & 0.692 & 0.559 & 0.633 & 0.656 \\
    \midrule[1.25pt]
    \rowcolor{whitegray} \multicolumn{7}{l}{\textit{Chemical LLMs (fine-tuning)}} \\
    \midrule
    MolT5-base & 0.540 & 0.457 & 0.634 & 0.485 & 0.578 & 0.569  \\
    \quad+\Algname & \textcolor{teal}{0.592}  & \textcolor{teal}{0.507}  & \textcolor{teal}{0.667} & \textcolor{teal}{0.523} & \textcolor{teal}{0.606} & \textcolor{teal}{0.619}  \\
    \midrule
    MolT5-large & 0.594 & 0.508 & 0.654 & 0.510 & 0.594 & 0.614 \\
    \quad +\Algname  & \textcolor{teal}{\textbf{0.645}} & \textcolor{teal}{\textbf{0.567}}  & \textcolor{teal}{\textbf{0.699}}  & \textcolor{teal}{\textbf{0.568}}  & \textcolor{teal}{\textbf{0.639}}& \textcolor{teal}{\textbf{0.666}} \\
    \midrule
    ChemT5-small & 0.553 & 0.462 & 0.633 & 0.481 & 0.574 & 0.583  \\
    \quad +\Algname & \textcolor{teal}{0.601}  & \textcolor{teal}{0.513} & \textcolor{teal}{0.664}  & \textcolor{teal}{0.519}  & \textcolor{teal}{0.603}  & \textcolor{teal}{0.624}  \\
    \midrule
    ChemT5-base & 0.580 & 0.490 & 0.647 & 0.498 & 0.586 & 0.604 \\
    \quad +\Algname & \textcolor{teal}{0.639}  & \textcolor{teal}{0.560} & \textcolor{teal}{0.687} & \textcolor{teal}{0.553} & \textcolor{teal}{0.626} & \textcolor{teal}{0.657}  \\
    \midrule[1.25pt]
    \rowcolor{whitegray} \multicolumn{7}{l}{\textit{General LLMs (without fine-tuning)}} \\
    \midrule
    Llama3 & 0.211 & 0.117 & 0.367 & 0.183 & 0.308 & 0.257  \\
    \quad +\Algname & \textcolor{teal}{0.259} & \textcolor{teal}{0.158} & \textcolor{teal}{0.401} & \textcolor{teal}{0.208} & \textcolor{teal}{0.324} & \textcolor{teal}{0.341}  \\
    \midrule
    GPT-4o & 0.232 & 0.128 & 0.389 & 0.183 & 0.307 & 0.291 \\ 
    \quad +\Algname & \textcolor{teal}{0.286} & \textcolor{teal}{0.174} & \textcolor{teal}{0.405} & \textcolor{teal}{0.199} & \textcolor{teal}{0.313} & \textcolor{teal}{0.341} \\
    \bottomrule[1.25pt]
    \end{tabular}
    }
    \vspace{-0.1in}
  \caption{\textbf{Molecule-to-text performance for ChEBI-20.}}\label{tab:mol2text}
\end{table}

In this section, we present our experiments on two frameworks: \molgiven and \molinfer. For the \molgiven framework, we consider molecule-to-text and retrosynthesis tasks. For the \molinfer framework, we address the text-to-molecule task. For clarity, in all tables, the \textcolor{teal}{teal} color indicates improvements over the vanilla model, and the best results are highlighted in \textbf{bold}.

\definecolor{whitegray}{RGB}{243, 243, 243} 

\begin{table}[t]
  \centering
  \resizebox{\linewidth}{!}{
\begin{tabular}{cccccccc}
    \toprule[1.25pt]
      & {\textbf{BL.2} } & {\textbf{BL.4} } &{\textbf{RO.1} } &{\textbf{RO.2} } &{\textbf{RO.L} } &{\textbf{ME.} } \\
    \midrule[1.25pt]
    \rowcolor{whitegray} \multicolumn{7}{l}{\textit{General LLM (without fine-tuning)}} \\
    \midrule
    \textcolor{black}{Mol-Instruct.} & 0.217 & 0.143 & 0.337 & 0.196 & 0.291 & 0.254 \\
    \textcolor{black}{\quad +\Algname} & \textbf{\textcolor{teal}{0.347}} & \textbf{\textcolor{teal}{0.275}}  & \textbf{\textcolor{teal}{0.601}} & \textbf{\textcolor{teal}{0.518}} & \textbf{\textcolor{teal}{0.593}} &\textbf{\textcolor{teal}{0.520}}  \\
    \bottomrule[1.25pt]
    \end{tabular}
    }
  \vspace{-0.1in}
  \caption{\textcolor{black}{\textbf{Molecule-to-text performance for Mol-Instructions.}}}\label{tab:mol2text_molins}
    \vspace{-0.15in}
\end{table}

\subsection{Analytic reasoning: Molecule-to-text}\label{subsec: mol_caption}

The molecule-to-text task aims to generate a precise and informative textual description that accurately represents the given molecule.

\begin{figure}[t]
    \centering
    \includegraphics[width=0.9\linewidth]{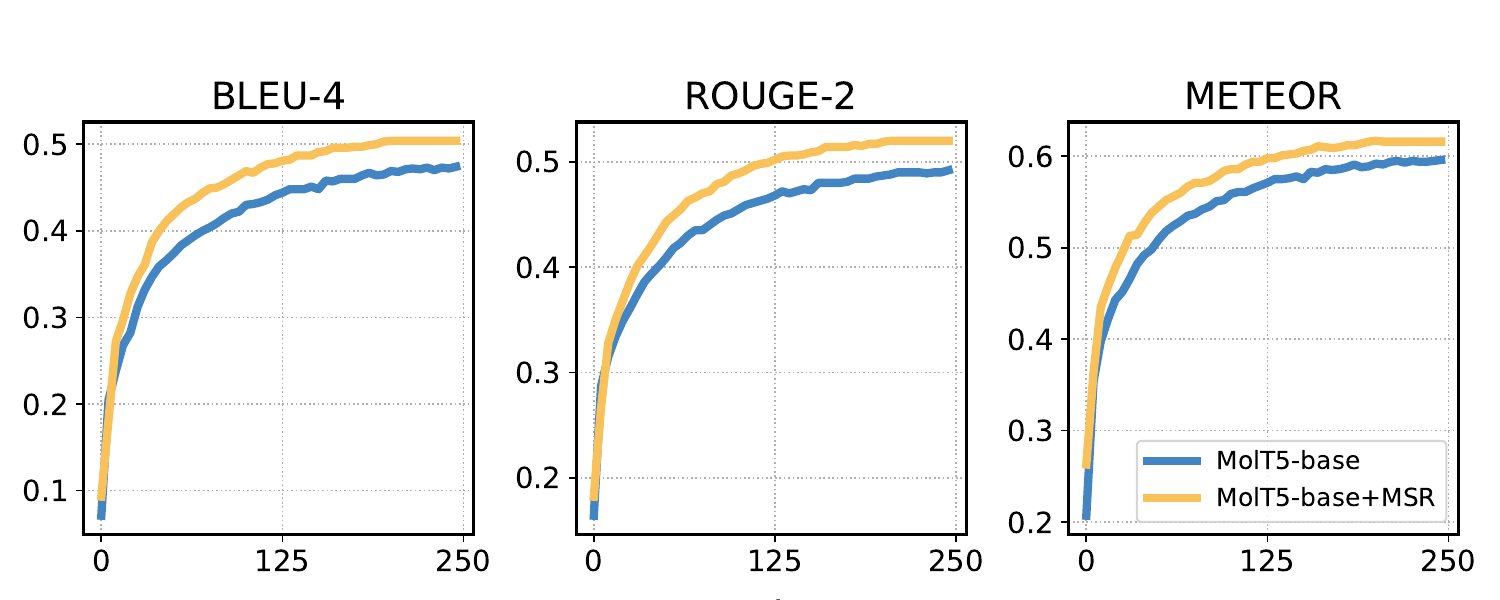}
    \vspace{-0.15in}
    \caption{\textbf{Fast performance improvement with \Algname.}}\label{fig: convergence}
    \vspace{-0.2in}
\end{figure}

\paragraph{Dataset.} We employ three datasets for the molecule-to-text task: (1) the recent L+M dataset ~\citep{edwards-etal-2024-lm}, \textcolor{black}{(2) the widely used ChEBI-20 dataset ~\citep{edwards-etal-2021-text2mol}, and the (3) Mol-instructions dataset ~\citep{fang2024molinstructions}. Each dataset consists of 182,331, 33,010, and 298,319 pairs of SMILES (or SELFIES) and their text descriptions, respectively. We use the same splits used in prior works.}

\paragraph{Baselines.} We evaluate the performance of \Algname with chemical and general LLMs. On the one hand, we employed two chemical LLMs: MolT5~\citep{edwards2022molt5} and Text+Chem T5~\citep[ChemT5;][]{christofidellis2023chemt5}. On the other hand, we employed \textcolor{black}{three} general LLMs: Llama3-8B-Instruct~\citep{touvron2023llamaopenefficientfoundation}, GPT-4o~\citep{openai2024gpt4technicalreport}\footnote{We used \texttt{gpt-4o-2024-05-13}.}, \textcolor{black}{and Mol-Instructions~\citep{fang2024molinstructions}}. Additionally, we include T5~\citep{raffel2020t5}, MolXPT~\citep{liu-etal-2023-molxpt}, BioT5~\citep{pei-etal-2023-biot5}, Meditron-7B~\citep{chen2023meditron70bscalingmedicalpretraining}, Mol2Lang-VLM~\citep{tran2024mollangvlm}, and BioT5+~\citep{pei2024enhanced} as baselines to compare absolute performance.

\paragraph{Experimental setup and metrics.} Chemical LLMs are trained following the process described in \cref{subsec: method_m2t}. For general LLMs without any domain-specific instruction tuning (Llama3 and GPT-4o), we cannot guarantee that the generated descriptions align with our training data. Therefore, we apply 10-shot in-context learning by attaching \Algname in the same manner as for chemical LLMs. \textcolor{black}{Additionally, for Mol-Instructions, we prompt with instructions enriched with \Algname. \textcolor{black}{Note that we additionally consider the molecular weight and IUPAC name components used by \citet{bran2024augmenting}, as they slightly improved the performance.}}

We evaluate the performance by comparing the generated description with the ground truth using six metrics: BLEU2, BLEU4~\citep{papineni2002bleu}, ROUGE1, ROUGE2, ROUGEL~\citep{banerjee2005meteor}, and METEOR~\citep{banerjee2005meteor}. We provide detailed experimental settings and prompts in \cref{appx: exp_m2t}.

\paragraph{Results.} We report the results in \cref{tab:mol2text_lm}, \cref{tab:mol2text}, and \cref{tab:mol2text_molins}. We observe that adding \Algname consistently improves performance across both chemical and general LLMs. Notably, in \cref{tab:mol2text}, ChemT5-base+\Algname achieves performance comparable to BioT5 (without MSR), despite BioT5 being pretrained on a larger dataset. \textcolor{black}{Furthermore, \cref{tab:mol2text_lm} shows that integrating \Algname with MolT5-base or MolT5-large yields superior performance compared to baseline models.} We provide examples of generated samples in \cref{fig: example} and \cref{fig: appx_m2t}. In addition, our method exhibits faster performance improvement, as illustrated in \cref{fig: convergence}.

\subsection{Analytic reasoning: Retrosynthesis}\label{subsec: retrosynthesis}

The retrosynthesis task aims to generate the corresponding set of reactant molecular representations based on a given product molecular representation.

\begin{table}[t]
  \centering
  \resizebox{\linewidth}{!}{
    \begin{tabular}{cccccccc}
    \toprule
     Models & \textbf{BL.} & \textbf{Ex.} & \textbf{Le.} $\downarrow$ & \textbf{MA.} & \textbf{RDK} & \textbf{Mo.} & \textbf{Val.} \\
    \midrule[1.25pt]
    \rowcolor{whitegray} \multicolumn{7}{l}{\textit{General LLM (without fine-tuning)}} \\
    \midrule
     Mol-Instruct. & \textbf{0.705} & 0.009  & 31.23 & 0.283 & 0.487 & 0.230 & \textbf{1.000} \\
     \quad + \Algname & 0.502 & \textcolor{teal}{\textbf{0.016}}  & \textcolor{teal}{\textbf{31.21}}& \textcolor{teal}{\textbf{0.315}} & \textcolor{teal}{\textbf{0.493}} & \textcolor{teal}{\textbf{0.273}} & \textbf{1.000}  \\
    \bottomrule
  \end{tabular}
  }
  \vspace{-0.1in}
    \caption{\textcolor{black}{\textbf{Retrosynthesis performance for Mol-Instructions.} BL., Ex., and Le. indicate BLEU, Exact, and Levenshtein distance. MA., RDK, and Mo. indicate MACCS, RDK, and Morgan fingerprint metrics. Val. indicates the validity.}} \label{tab: retro}
 \vspace{-0.2in}
\end{table}

\textcolor{black}{\paragraph{Dataset and baselines.} We employ the dataset and the model used by Mol-instructions ~\citep{fang2024molinstructions}. The dataset consists of 129,684 product and reactant pairs.}

\textcolor{black}{\paragraph{Experimental setup and metrics.} As the input molecule (i.e., product) is given for the retrosynthesis task, we follow the framework proposed in \cref{subsec: method_m2t}. The performance is evaluated by comparing the generated molecules with the ground truth with eight metrics: SMILES comparison metrics (BLEU, Exact, and Levenshtein distance \citep{miller2009levenshtein}), fingerprint similarity metrics (MACCS FTS \citep{durant2002maccs}, RDK FTS \citep{schneider2015rdk}, and Morgan FTS \citep{rogers2010morgan}), a molecular distribution metric (Fréchet ChemNet Distance (FCD) \citep{preuer2018fcd}), and the validity of the molecule.}

\textcolor{black}{\paragraph{Results.} We report the results in \cref{tab: retro}, showing that incorporating \Algname improves performance across all metrics except BLEU. This highlights its effectiveness in enhancing complex tasks. Notably, while we report BLEU for consistency with prior work, it is less critical than other metrics, as it evaluates string-based accuracy rather than molecular structure alignment.}

\subsection{Synthetic reasoning: Text-to-molecule}\label{subsec: mol_gen}

The text-to-molecule task is the inverse of molecule-to-text, aiming to generate a molecular representation based on a given textual description.

\textcolor{black}{\paragraph{Dataset.} We employ two datasets for the text-to-molecule task: (1) L+M ~\citep{edwards-etal-2024-lm} and (2) ChEBI-20 ~\citep{edwards-etal-2021-text2mol}. We followed the same settings used in \cref{subsec: mol_caption}.}

\definecolor{whitegray}{RGB}{243, 243, 243} 
\begin{table}[t]
  \centering
  \resizebox{\linewidth}{!}{
\begin{tabular}{ccccccccc}
    \toprule[1.25pt]
    Models & \textbf{Fo.} & \textbf{Ch.} & \textbf{Ar.} & \textbf{Ri.} & \textbf{Fu.} & \textbf{Ch.} & \textbf{We.} & \textbf{Na.} \\
    \midrule[1.25pt]
 \rowcolor{whitegray} \multicolumn{9}{l}{\textit{Chemical LLMs (\Algname fine-tuning) - L+M}} \\
 \midrule
 MolT5-small & 0.048 & 0.235 & 0.783 & 0.781 & 0.849 & 0.647 & 0.418 & 0.248 \\
\midrule
 MolT5-base & 0.426 & 0.527 & 0.825 & 0.813 & 0.889 & 0.807 & 0.615 & 0.309 \\
 \midrule
 MolT5-large & 0.221 & 0.317 & 0.820 & 0.809 & 0.872 & 0.691 & 0.529 & 0.576  \\
 \midrule[1.25pt]
    \rowcolor{whitegray} \multicolumn{9}{l}{\textit{Chemical LLMs (\Algname fine-tuning) - MolT5}} \\
    \midrule
    MolT5-base & 0.458 & 0.922 & 0.926 & 0.930 & 0.957 & 0.798 & 0.606 & 0.512\\
    \midrule
    ChemT5-small & 0.447 & 0.920 & 0.930 & 0.926 & 0.954 & 0.788 & 0.634 & 0.495 \\
    \midrule
    ChemT5-base & 0.475 & 0.925 & 0.931 & 0.930 & 0.960 & 0.799 & 0.641 & 0.525 \\
    \midrule[1.25pt]
    \rowcolor{whitegray} \multicolumn{9}{l}{\textit{General LLMs (\Algname few-shot learning) - MolT5}} \\
    \midrule
    Llama3 & 0.084 & 0.174 & 0.593 & 0.362 & 0.137 & 0.450 & 0.435 & 0.015\\
    \midrule
    GPT-4o & 0.298 & 0.235 & 0.718 & 0.464 & 0.298 & 0.485 & 0.728 & 0.040 \\
    \bottomrule[1.25pt]
    \end{tabular}
    }
     \vspace{-0.1in}
  \caption{\textbf{Reasoning accuracy for each structural information.} Fo., Ch., Ar., Ri., Fu., Ch., We., Na., stand for molecular formula, longest carbon chain length, aromatic rings, ring compounds, functional groups, chiarlity, molecular weight, and IUPAC name, respectively.   \label{tab:reason}}
    \vspace{-0.2in}
\end{table}

\textcolor{black}{\paragraph{Baselines.} Two popular chemical LLMs, including MolT5~\citep{edwards2022molt5} and ChemT5~\citep{christofidellis2023chemt5}, serve as our baselines. Notably, we exclude general LLMs from this evaluation due to their insufficient reasoning accuracy as shown in \cref{tab:reason}. In detail, their low accuracy implies that their reasoning cannot guide the answer appropriately, even in a few-shot learning setting. For completeness, we provide the results for general LLMs in \cref{appx: samp_t2m}. Additional baselines are consistent with those in \cref{subsec: mol_caption} other than Lang2Mol-Diff~\citep{nguyen2024langmoldiff}.}

\paragraph{Experimental setup and metrics.} We follow the framework proposed in \cref{subsec: method_t2m}. We provide detailed experimental settings and prompts in \cref{appx: exp_t2m}. The performance is evaluated using the same metrics described in \cref{subsec: retrosynthesis}.

\definecolor{whitegray}{RGB}{243, 243, 243}

\begin{table}[t]
  \vspace{-0.1in}
  \centering
  \resizebox{\linewidth}{!}{
\begin{tabular}{ccccccccc}
    \toprule[1.25pt]
      & {\textbf{BL.} } & {\textbf{Ex.} } &{\textbf{Le.} $\downarrow$} &{\textbf{MA.}} &{\textbf{RDK}} &{\textbf{Mo.}} &{\textbf{FCD}$\downarrow$} &{\textbf{Val.}}\\
      \midrule[1.25pt]
       \rowcolor{whitegray} \multicolumn{9}{l}{\textit{Baselines (without reasoning)}} \\
    \midrule
    Meditron-7B & 0.694 & 0.010 & 46.49 & 0.772 & 0.693 & 0.501 & \phantom{0}2.46 & 0.996 \\
    \midrule
    Lang2Mol-Diff & 0.543 & 0.000 & 55.87 & 0.606 & 0.332 & 0.328 & 38.09 & \textbf{1.000} \\
    
    \midrule
    BioT5+ & \textbf{0.731} & 0.010 & 41.47 & 0.781 & 0.709 & 0.515 & \phantom{0}3.29 & \textbf{1.000} \\
    \midrule[1.25pt]
    \rowcolor{whitegray} \multicolumn{9}{l}{\textit{Chemical LLMs (fine-tuning)}} \\
    \midrule
    MolT5-small & 0.566 & 0.000 & 56.34 & 0.642 & 0.581 & 0.374 & NaN & 0.805 \\
    \quad +\Algname & \textcolor{teal}{0.730} & \textcolor{teal}{0.002} &  \textcolor{teal}{41.15} &  \textcolor{teal}{0.798} &  \textcolor{teal}{0.712} & \textcolor{teal}{0.514} & \textcolor{teal}{\phantom{0}2.82} & \textcolor{teal}{0.995} \\
    \midrule
    MolT5-base & 0.684 & 0.000 & 44.79 & 0.760 & 0.652 & 0.475 & NaN & \textbf{1.000} \\
    \quad +\Algname & \textcolor{teal}{0.706} & \textcolor{teal}{0.052} & \textcolor{teal}{40.18} & \textcolor{teal}{0.825} & \textcolor{teal}{0.762} & \textcolor{teal}{0.548} & \textcolor{teal}{\textbf{\phantom{0}1.45}} & 0.997 \\
    \midrule
    MolT5-large & 0.564 & 0.000 & 55.40 & 0.757 & 0.650 & 0.395 & 17.50 & 0.994  \\
    \quad +\Algname & \textcolor{teal}{0.710} & \textcolor{teal}{\textbf{0.111}} & \textcolor{teal}{\textbf{39.54}} & \textcolor{teal}{\textbf{0.837}} & \textcolor{teal}{\textbf{0.783}} & \textcolor{teal}{\textbf{0.560}} & \textcolor{teal}{\phantom{0}1.54} & \textcolor{teal}{0.999} \\
    \bottomrule[1.25pt]
    \end{tabular}
    }
    \vspace{-0.1in}
  \caption{\textcolor{black}{\textbf{Text-to-molecule performance for L+M val.}}}\label{tab:text2mol_lm}
\end{table}

\paragraph{Reasoning accuracy.} We first measure the reasoning accuracy to filter out low-accuracy components that may misguide the answer. The detailed computation process is in \cref{appx: exp_t2m}. The reasoning accuracies are provided in \cref{tab:reason}. Our results show that our fine-tuned reasoning modules exhibit superior accuracy compared to larger general LLMs, underscoring their ability to understand molecular structures effectively. However, they still struggle with certain structural elements, such as molecular formula, molecular weight, and IUPAC name, with additional challenges in carbon chain length and chirality in the L+M dataset. Consequently, we exclude these components.

\paragraph{Results.} The results are reported in \cref{tab:text2mol_lm} and \cref{tab:text2mol}. Incorporating \Algname into the molecular description always improved performance. In particular, integrating \Algname into the ChemT5-base achieves state-of-the-art performance compared to the recent baselines, validating its efficacy. Surprisingly, our \Algname even improves the performance of smaller models beyond that of the vanilla larger models, e.g., MolT5-base+\Algname showed superior performance to MolT5-large. \textcolor{black}{We provide examples of generated samples in \cref{appx: samp_m2t}.}

\definecolor{whitegray}{RGB}{243, 243, 243}

\begin{table}[t]
  \vspace{-0.1in}
  \centering
  \resizebox{\linewidth}{!}{
\begin{tabular}{ccccccccc}
    \toprule[1.25pt]
      & {\textbf{BL.} } & {\textbf{Ex.} } &{\textbf{Le.} $\downarrow$} &{\textbf{MA.}} &{\textbf{RDK}} &{\textbf{Mo.}} &{\textbf{FCD}$\downarrow$} &{\textbf{Val.}}\\
      \midrule[1.25pt]
       \rowcolor{whitegray} \multicolumn{9}{l}{\textit{Baselines (without reasoning)}} \\
    \midrule
    T5-base & 0.762 & 0.069 & 24.95 & 0.731 & 0.605 & 0.545 & 2.48 & 0.660 \\
    \midrule
    MolXPT & - &  0.215 & - & 0.859 & 0.757 & 0.667 & 0.45 &  0.983 \\
    \midrule
    BioT5 & 0.867 & 0.413 & 15.10 & 0.886 & 0.801 & 0.734 & 0.43 & \textbf{1.000}  \\
    \midrule[1.25pt]
    \rowcolor{whitegray} \multicolumn{9}{l}{\textit{Chemical LLMs (fine-tuning)}} \\
    \midrule
    MolT5-base & 0.769 & 0.081 & 24.46 & 0.721 & 0.588 & 0.529 & 2.18 & 0.772  \\
    \quad +\Algname & \textcolor{teal}{0.863} & \textcolor{teal}{0.385} & \textcolor{teal}{13.91} & \textcolor{teal}{0.918} & \textcolor{teal}{0.843} & \textcolor{teal}{0.783} & \textcolor{teal}{0.29} & \textcolor{teal}{0.983} \\
    \midrule
    MolT5-large & 0.854 & 0.311 & 16.07 & 0.834 & 0.746 & 0.684 & 1.20 & 0.905 \\
    \quad +\Algname & \textcolor{teal}{\textbf{0.886}} & \textcolor{teal}{0.391} & \textcolor{teal}{12.98} & \textcolor{teal}{0.906} & \textcolor{teal}{0.822} & \textcolor{teal}{0.765} & \textcolor{teal}{0.35} & \textcolor{teal}{0.947} \\
    \midrule
    ChemT5-small & 0.739 & 0.157 & 28.54 & 0.859 & 0.736 & 0.660 & 0.07 & 0.776 \\
    \quad +\Algname & \textcolor{teal}{0.874} & \textcolor{teal}{0.381} & \textcolor{teal}{13.22} & \textcolor{teal}{0.918} & \textcolor{teal}{0.845} & \textcolor{teal}{0.787} & 0.29 & \textcolor{teal}{0.976} \\
    \midrule
    ChemT5-base & 0.750 & 0.212 & 27.39 & 0.874 & 0.767 & 0.697 & \textbf{0.06} & 0.792 \\
    \quad +\Algname & \textcolor{teal}{0.878} & \textcolor{teal}{\textbf{0.421}} & \textcolor{teal}{\textbf{12.76}} & \textcolor{teal}{\textbf{0.924}} & \textcolor{teal}{\textbf{0.856}} & \textcolor{teal}{\textbf{0.804}} & 0.26 & \textcolor{teal}{0.982} \\
    \bottomrule[1.25pt]
    \end{tabular}
    }
 \vspace{-0.1in}
  \caption{\textbf{Text-to-molecule performance for ChEBI-20.}}\label{tab:text2mol}
    \vspace{-0.2in}
\end{table}

\subsection{Ablation study}\label{subsec: ablation}

We perform ablation studies on matching ratio-based rejection sampling and each structural component. Here, we utilize ChemT5-small on the ChEBI-20 dataset. \textcolor{black}{Due to limited space, additional ablation study results, including a comparison with ChemCrow \citep{bran2024augmenting}, prior work on the reasoning for chemistry tasks, and extra structural component, are provided in \cref{appx: samp_ab}.}

    

\begin{figure}[t]
    \centering
\includegraphics[width=0.98\linewidth]{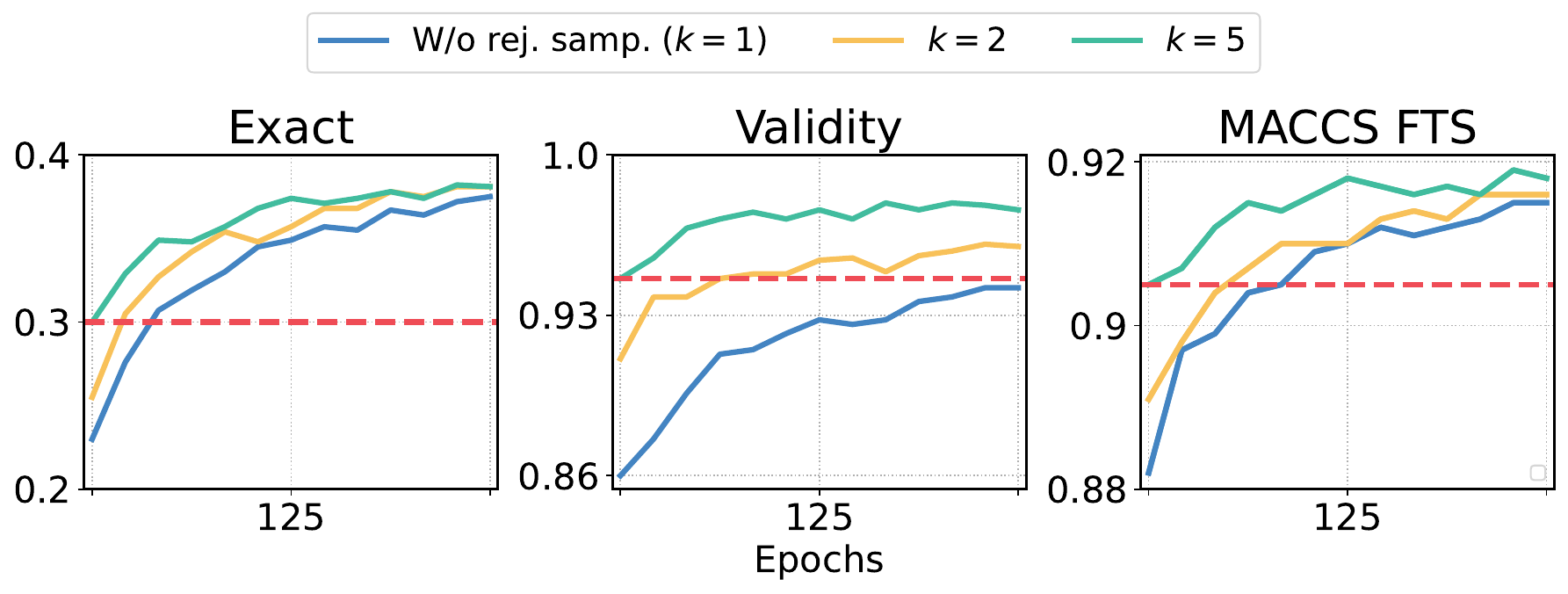}
    \vspace{-0.15in}    \caption{\textcolor{black}{\textbf{Impact of $k$ in rejection sampling.} Dotted lines indicate the initial performance of $k=5$.}}\label{fig: iter}
    \vspace{-0.2in}
\end{figure}

         

\paragraph{Matching ratio-based rejection sampling.} We discuss the efficacy of matching ratio-based rejection sampling \textcolor{black}{and the impact of the number of samples $k$} in text-to-molecule. We compare the results of without ($k=1$) and with the rejection sampling \textcolor{black}{($k\in\{2,5\}$)}. As demonstrated in \cref{fig: iter}, the rejection sampling improves performance by encouraging the output to follow the \Algname. Notably, increasing $k$ beyond 5 does not further improve performance, implying that $k=5$ is sufficient.

\begin{figure}[t]
\vspace{-0.1in}  
    \centering\includegraphics[width=\linewidth]{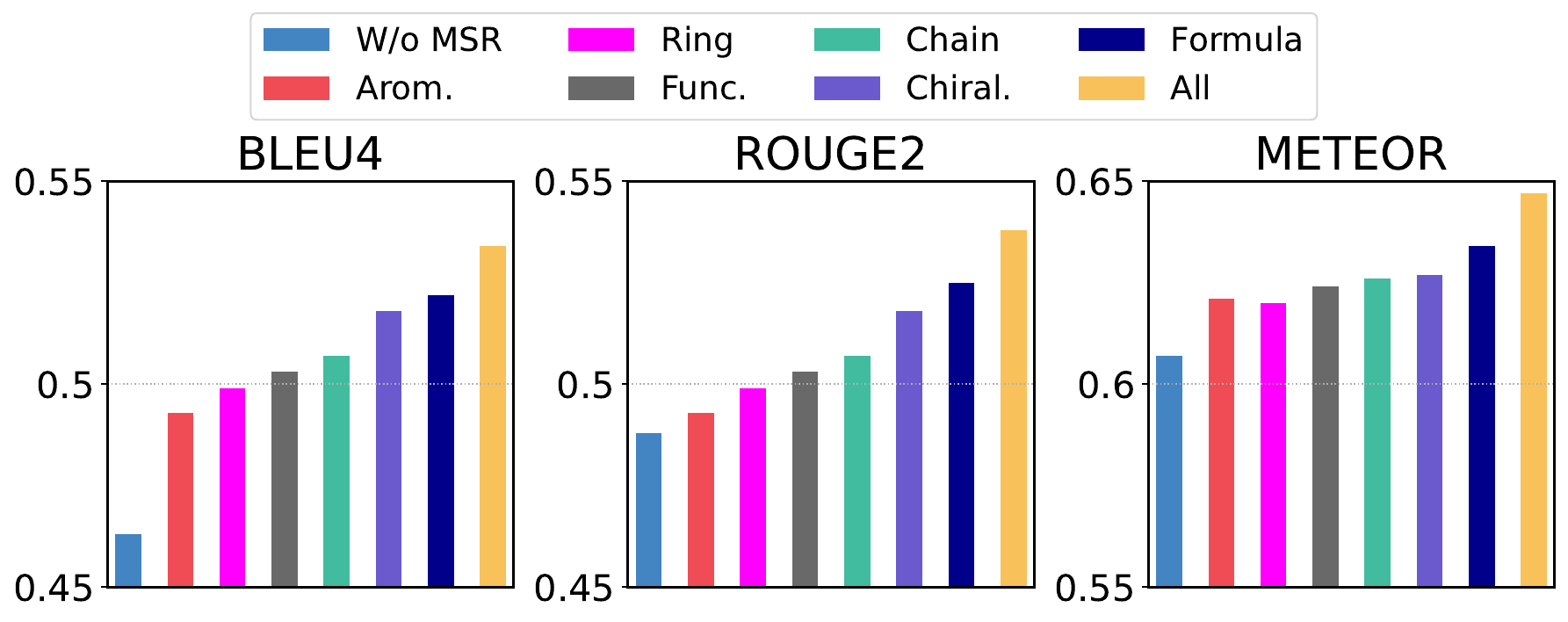}
    \vspace{-0.3in}
    \caption{\textcolor{black}{\textbf{Impact of each structural component.}}}
    \label{fig: appx_ablation}
    \vspace{-0.2in}
\end{figure}

\textcolor{black}{\paragraph{Structural component.} To verify the effectiveness of each component, we evaluated the performance of molecule-to-text using each structural information component individually. We provide the results in \cref{fig: appx_ablation}. Incorporating each single component resulted in better performance compared to the baseline model without any reasoning. Notably, combining all the proposed structural elements yielded the best results, validating the effectiveness of our comprehensive approach.}

\paragraph{Additional molecular descriptors.} In addition to the proposed six structural components, we conducted experiments using three more advanced molecular descriptors: the Morgan fingerprint and two electronic properties—topological polar surface area (TPSA) and molar refractivity (MR). Specifically, the Morgan fingerprint encodes local substructures within a specified radius; TPSA represents the sum of the surface areas of all polar atoms and their attached hydrogen atoms; and MR quantifies the total polarizability of a molecule.

To verify the effectiveness of each additional descriptor, we evaluated the performance of molecule captioning using ChemT5-small. We provide the results in \cref{fig: appx_add}. We observed that incorporating all three additional descriptors together did not further improve the performance of \Algname, although applying each additional descriptor individually improved performance. This validates the importance of structural information and the sufficiency of our proposed structural components.

\section{Related work}
\paragraph{Large language models for chemistry.} 
General LLMs often struggle to solve basic chemistry problems and molecular tasks~\citep{white2023assessment, nascimento2023understandchemistry, guo2023can}. To address this issue, prior works have introduced chemical LLMs by pre-training models on molecule-related texts~\citep{edwards2022molt5, christofidellis23unifying, liu-etal-2023-molxpt, pei-etal-2023-biot5}, through instruction tuning~\citep{fang2024molinstructions, cao2023instructmol}, and using retrieval-based in-context learning~\citep{Li2024molregpt}. Our work focuses on reasoning processes that are broadly applicable to these chemical and general LLMs.

\textcolor{black}{\paragraph{Reasoning of LLMs.} Generating intermediate reasoning before arriving at a final answer~\citep{wei2023chainofthoughtpromptingelicitsreasoning, kojima2022large} improves the overall quality of generated answers. However, the ability to perform complex reasoning remains limited to huge models ($>$100B parameters).}

\begin{figure}[t]
\centering\includegraphics[width=\linewidth]{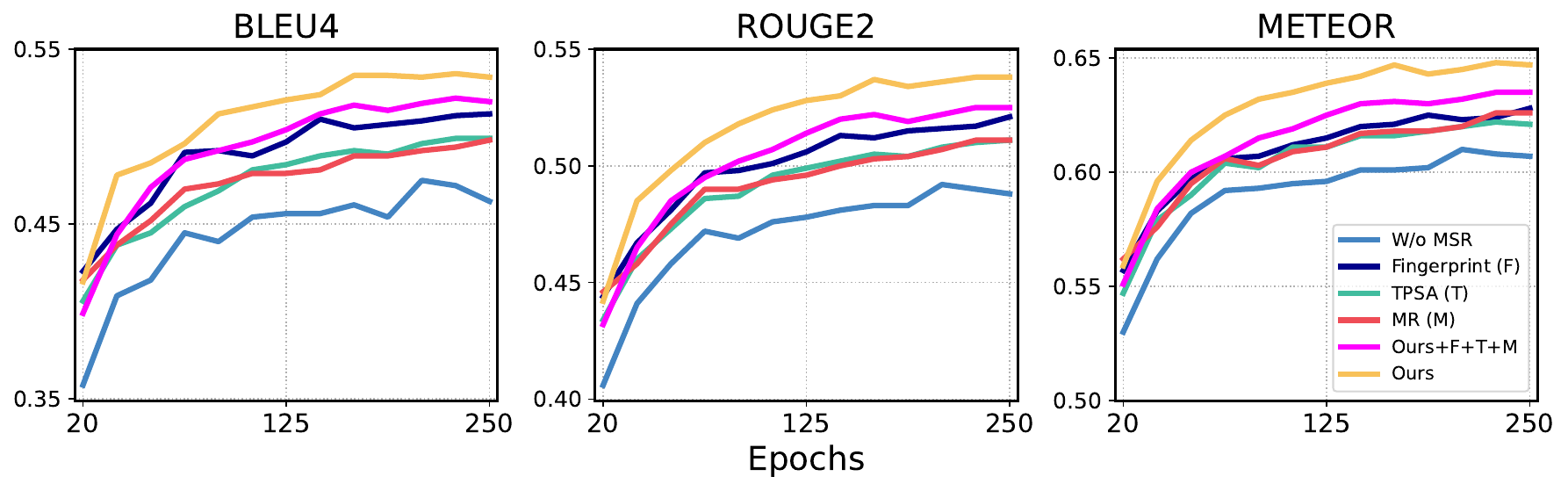}
    \caption{\textbf{The impact of additional molecular descriptors.}}
    \label{fig: appx_add}
    \vspace{-0.2in}
\end{figure}

To address this challenge, various approaches have been introduced to distill knowledge from larger language models to smaller ones ($<$10B). Specifically, \citet{ho-etal-2023-large, fu2023specializing, magister-etal-2023-teaching} employed the larger models as teacher models to generate rationales for fine-tuning smaller student models. Nevertheless, even recent LLMs struggle to generate appropriate rationales that demonstrate a correct understanding of molecular structures (as described in \cref{fig: failure_case} and \cref{subsec: failure}), restricting the efficacy of LLMs in generating rationales for molecular tasks.

\paragraph{Reasoning for chemistry.} Recently, a few works have extended the reasoning of LLMs to address chemistry problems. For instance, \citet{ouyang2024structuredchemistryreasoninglarge} proposed employing the program-of-thoughts~\citep[PoT;][]{chen2023program} to handle chemical question-answering tasks. Additionally, \citet{jin2024prollm} presented the protein chain-of-thought (ProCoT) to replicate the signaling pathways in the protein-protein interaction (PPI) problem. \textcolor{black}{Despite these advances, none of these works are generally applicable to various molecular tasks.} We note that \citet{bran2024augmenting} provided a reasoning approach comparable to ours, but their rationales are less focused on molecular structures, e.g., rationales based on tools like \textit{LitSearch/WebSearch}, \textit{PatentCheck}, \textit{ReactionPlanner}, and \textit{SMILES2Price}. \textcolor{black}{
Moreover, it shows limited performance improvement in molecule generation and molecule captioning tasks, as observed in \cref{appx: samp_ab}.}

\section{Conclusion}

We introduced \Algname, a molecular structural reasoning framework that enhances LLMs’ understanding of molecules by explicitly incorporating key structural features. Our investigation revealed recent LLMs’ limitations in inferring structural information, emphasizing the need for explicit reasoning. Fine-tuning chemical LLMs with \Algname led to consistent improvements across three molecular tasks, highlighting the effectiveness of domain-specific models for molecular reasoning.

\clearpage

\section*{Broader impacts}

Our work contributes to the development of more interpretable and reliable models for molecular applications. By incorporating explicit molecular reasoning, our framework has the potential to enhance molecular understanding and improve decision-making in areas such as drug discovery, materials science, and chemical synthesis. However, as with any AI-driven molecular generation system, there are potential risks and ethical concerns. For instance, the generation of harmful or toxic compounds poses significant safety challenges. Additionally, over-reliance on AI-generated molecular reasoning without expert validation could lead to unintended consequences in scientific and industrial applications.

\section*{Limitations}
One limitation of \Algname is its reliance on the accuracy of structural information in \molinfer. While external tools like RDKit provide precise structural information for molecule-forward reasoning, errors in molecule-backward reasoning (where structural features must be inferred) could degrade performance. However, appropriate filtering based on reasoning accuracy can prevent this to some extent. Additionally, we assume that the given molecular representations are accurate when we extract the structural information. However, real-world data can be noisy or incomplete. Extending \Algname to handle uncertain molecular inputs via self-correction remains an open challenge. 

\section*{Reproducibility} All experimental code related to this paper is available at \url{https://github.com/yunhuijang/MSR}. Detailed insights regarding the experiments, encompassing dataset and model specifics, are available in \cref{sec: exp}. For intricate details like hyperparameters, consult \cref{appx: exp}.

\section*{Acknowledgement}

This work was partly supported by Institute for Information \& communications Technology Planning \& Evaluation(IITP) grant funded by the Korea government(MSIT) (RS-2019-
II190075, Artificial Intelligence Graduate School Support Program(KAIST)), National Research Foundation of Korea(NRF) grant funded by the Ministry of Science and ICT(MSIT) (No. RS-2022-
NR072184), GRDC(Global Research Development Center) Cooperative Hub Program through
the National Research Foundation of Korea(NRF) grant funded by the Ministry of Science and
ICT(MSIT) (No. RS-2024-00436165), and the Institute of Information \& Communications Technology Planning \& Evaluation(IITP) grant funded by the Korea government(MSIT) (RS-2025-02304967, AI Star Fellowship(KAIST)).

\clearpage
\newpage



\bibliography{custom}

\begin{thebibliography}{45}
\providecommand{\natexlab}[1]{#1}

\bibitem[{Banerjee and Lavie(2005)}]{banerjee2005meteor}
Satanjeev Banerjee and Alon Lavie. 2005.
\newblock Meteor: An automatic metric for mt evaluation with improved correlation with human judgments.
\newblock In \emph{Proceedings of the acl workshop on intrinsic and extrinsic evaluation measures for machine translation and/or summarization}, pages 65--72.

\bibitem[{Cao et~al.(2023)Cao, Liu, Lu, Yao, and Li}]{cao2023instructmol}
He~Cao, Zijing Liu, Xingyu Lu, Yuan Yao, and Yu~Li. 2023.
\newblock \href {https://arxiv.org/abs/2311.16208} {Instructmol: Multi-modal integration for building a versatile and reliable molecular assistant in drug discovery}.
\newblock \emph{Preprint}, arXiv:2311.16208.

\bibitem[{Castro~Nascimento and Pimentel(2023)}]{nascimento2023understandchemistry}
Cayque~Monteiro Castro~Nascimento and Andr{\'e}Silva Pimentel. 2023.
\newblock \href {https://doi.org/10.1021/acs.jcim.3c00285} {Do large language models understand chemistry? a conversation with chatgpt}.
\newblock \emph{Journal of Chemical Information and Modeling}, 63(6):1649--1655.

\bibitem[{Chen et~al.(2023{\natexlab{a}})Chen, Ma, Wang, and Cohen}]{chen2023program}
Wenhu Chen, Xueguang Ma, Xinyi Wang, and William~W. Cohen. 2023{\natexlab{a}}.
\newblock \href {https://openreview.net/forum?id=YfZ4ZPt8zd} {Program of thoughts prompting: Disentangling computation from reasoning for numerical reasoning tasks}.
\newblock \emph{Transactions on Machine Learning Research}.

\bibitem[{Chen et~al.(2023{\natexlab{b}})Chen, Cano, Romanou, Bonnet, Matoba, Salvi, Pagliardini, Fan, Köpf, Mohtashami, Sallinen, Sakhaeirad, Swamy, Krawczuk, Bayazit, Marmet, Montariol, Hartley, Jaggi, and Bosselut}]{chen2023meditron70bscalingmedicalpretraining}
Zeming Chen, Alejandro~Hernández Cano, Angelika Romanou, Antoine Bonnet, Kyle Matoba, Francesco Salvi, Matteo Pagliardini, Simin Fan, Andreas Köpf, Amirkeivan Mohtashami, Alexandre Sallinen, Alireza Sakhaeirad, Vinitra Swamy, Igor Krawczuk, Deniz Bayazit, Axel Marmet, Syrielle Montariol, Mary-Anne Hartley, Martin Jaggi, and Antoine Bosselut. 2023{\natexlab{b}}.
\newblock \href {https://arxiv.org/abs/2311.16079} {Meditron-70b: Scaling medical pretraining for large language models}.
\newblock \emph{Preprint}, arXiv:2311.16079.

\bibitem[{Christofidellis et~al.(2023{\natexlab{a}})Christofidellis, Giannone, Born, Winther, Laino, and Manica}]{christofidellis2023chemt5}
Dimitrios Christofidellis, Giorgio Giannone, Jannis Born, Ole Winther, Teodoro Laino, and Matteo Manica. 2023{\natexlab{a}}.
\newblock \href {https://proceedings.mlr.press/v202/christofidellis23a.html} {Unifying molecular and textual representations via multi-task language modelling}.
\newblock In \emph{Proceedings of the 40th International Conference on Machine Learning}, volume 202 of \emph{Proceedings of Machine Learning Research}, pages 6140--6157. PMLR.

\bibitem[{Christofidellis et~al.(2023{\natexlab{b}})Christofidellis, Giannone, Born, Winther, Laino, and Manica}]{christofidellis23unifying}
Dimitrios Christofidellis, Giorgio Giannone, Jannis Born, Ole Winther, Teodoro Laino, and Matteo Manica. 2023{\natexlab{b}}.
\newblock \href {https://proceedings.mlr.press/v202/christofidellis23a.html} {Unifying molecular and textual representations via multi-task language modelling}.
\newblock In \emph{Proceedings of the 40th International Conference on Machine Learning}, volume 202 of \emph{Proceedings of Machine Learning Research}, pages 6140--6157. PMLR.

\bibitem[{Durant et~al.(2002)Durant, Leland, Henry, and Nourse}]{durant2002maccs}
Joseph~L Durant, Burton~A Leland, Douglas~R Henry, and James~G Nourse. 2002.
\newblock Reoptimization of mdl keys for use in drug discovery.
\newblock \emph{Journal of chemical information and computer sciences}, 42(6):1273--1280.

\bibitem[{Edwards et~al.(2022)Edwards, Lai, Ros, Honke, Cho, and Ji}]{edwards2022molt5}
Carl Edwards, Tuan Lai, Kevin Ros, Garrett Honke, Kyunghyun Cho, and Heng Ji. 2022.
\newblock \href {https://doi.org/10.18653/v1/2022.emnlp-main.26} {Translation between molecules and natural language}.
\newblock In \emph{Proceedings of the 2022 Conference on Empirical Methods in Natural Language Processing}, pages 375--413, Abu Dhabi, United Arab Emirates. Association for Computational Linguistics.

\bibitem[{Edwards et~al.(2024)Edwards, Wang, Zhao, and Ji}]{edwards-etal-2024-lm}
Carl Edwards, Qingyun Wang, Lawrence Zhao, and Heng Ji. 2024.
\newblock \href {https://doi.org/10.18653/v1/2024.langmol-1.1} {{L}+{M}-24: Building a dataset for {L}anguage+{M}olecules @ {ACL} 2024}.
\newblock In \emph{Proceedings of the 1st Workshop on Language + Molecules (L+M 2024)}, pages 1--9, Bangkok, Thailand. Association for Computational Linguistics.

\bibitem[{Edwards et~al.(2021)Edwards, Zhai, and Ji}]{edwards-etal-2021-text2mol}
Carl Edwards, ChengXiang Zhai, and Heng Ji. 2021.
\newblock \href {https://doi.org/10.18653/v1/2021.emnlp-main.47} {{T}ext2{M}ol: Cross-modal molecule retrieval with natural language queries}.
\newblock In \emph{Proceedings of the 2021 Conference on Empirical Methods in Natural Language Processing}, pages 595--607, Online and Punta Cana, Dominican Republic. Association for Computational Linguistics.

\bibitem[{Fang et~al.(2024)Fang, Liang, Zhang, Liu, Huang, Chen, Fan, and Chen}]{fang2024molinstructions}
Yin Fang, Xiaozhuan Liang, Ningyu Zhang, Kangwei Liu, Rui Huang, Zhuo Chen, Xiaohui Fan, and Huajun Chen. 2024.
\newblock \href {https://openreview.net/forum?id=Tlsdsb6l9n} {Mol-instructions: A large-scale biomolecular instruction dataset for large language models}.
\newblock In \emph{The Twelfth International Conference on Learning Representations}.

\bibitem[{Fu et~al.(2023)Fu, Peng, Ou, Sabharwal, and Khot}]{fu2023specializing}
Yao Fu, Hao Peng, Litu Ou, Ashish Sabharwal, and Tushar Khot. 2023.
\newblock \href {https://proceedings.mlr.press/v202/fu23d.html} {Specializing smaller language models towards multi-step reasoning}.
\newblock In \emph{Proceedings of the 40th International Conference on Machine Learning}, volume 202 of \emph{Proceedings of Machine Learning Research}, pages 10421--10430. PMLR.

\bibitem[{Ganeeva et~al.(2024)Ganeeva, Sakhovskiy, Khrabrov, Savchenko, Kadurin, and Tutubalina}]{ganeeva2024lost}
Veronika Ganeeva, Andrey Sakhovskiy, Kuzma Khrabrov, Andrey Savchenko, Artur Kadurin, and Elena Tutubalina. 2024.
\newblock Lost in translation: Chemical language models and the misunderstanding of molecule structures.
\newblock In \emph{Findings of the Association for Computational Linguistics: EMNLP 2024}, pages 12994--13013.

\bibitem[{Guo et~al.(2023)Guo, Nan, Liang, Guo, Chawla, Wiest, Zhang et~al.}]{guo2023can}
Taicheng Guo, Bozhao Nan, Zhenwen Liang, Zhichun Guo, Nitesh Chawla, Olaf Wiest, Xiangliang Zhang, et~al. 2023.
\newblock What can large language models do in chemistry? a comprehensive benchmark on eight tasks.
\newblock \emph{Advances in Neural Information Processing Systems}, 36:59662--59688.

\bibitem[{Hansch et~al.(2000)Hansch, Mckarns, Smith, and Doolittle}]{hansch2000phenol}
Corwin Hansch, Susan Mckarns, Carr Smith, and David Doolittle. 2000.
\newblock \href {https://doi.org/10.1016/S0009-2797(00)00171-X} {Comparative qsar evidence for a free-radical mechanism of phenol-induced toxicity}.
\newblock \emph{Chemico-Biological Interactions}, 127:61--72.

\bibitem[{Ho et~al.(2023)Ho, Schmid, and Yun}]{ho-etal-2023-large}
Namgyu Ho, Laura Schmid, and Se-Young Yun. 2023.
\newblock \href {https://doi.org/10.18653/v1/2023.acl-long.830} {Large language models are reasoning teachers}.
\newblock In \emph{Proceedings of the 61st Annual Meeting of the Association for Computational Linguistics (Volume 1: Long Papers)}, pages 14852--14882, Toronto, Canada. Association for Computational Linguistics.

\bibitem[{Jin et~al.(2024)Jin, Xue, Wang, Kang, Ye, Zhou, Du, and Zhang}]{jin2024prollm}
Mingyu Jin, Haochen Xue, Zhenting Wang, Boming Kang, Ruosong Ye, Kaixiong Zhou, Mengnan Du, and Yongfeng Zhang. 2024.
\newblock \href {https://openreview.net/forum?id=2nTzomzjjb} {Pro{LLM}: Protein chain-of-thoughts enhanced {LLM} for protein-protein interaction prediction}.
\newblock In \emph{First Conference on Language Modeling}.

\bibitem[{Kant(1899)}]{critique1899kant}
I.~Kant. 1899.
\newblock \href {https://doi.org/10.1037/11654-000} {\emph{Critique of pure reason}}.

\bibitem[{Kojima et~al.(2022)Kojima, Gu, Reid, Matsuo, and Iwasawa}]{kojima2022large}
Takeshi Kojima, Shixiang~Shane Gu, Machel Reid, Yutaka Matsuo, and Yusuke Iwasawa. 2022.
\newblock \href {https://openreview.net/forum?id=e2TBb5y0yFf} {Large language models are zero-shot reasoners}.
\newblock In \emph{Advances in Neural Information Processing Systems}.

\bibitem[{Landrum et~al.(2024)Landrum, Tosco, Kelley, Rodriguez, Cosgrove, Vianello, sriniker, Gedeck, Jones, NadineSchneider, Kawashima, Nealschneider, Dalke, Swain, Cole, Turk, Savelev, Vaucher, Wójcikowski, Take, Scalfani, Walker, Probst, Ujihara, Pahl, guillaume godin, Lehtivarjo, tadhurst cdd, Bérenger, and Bisson}]{greg2024rdkit}
Greg Landrum, Paolo Tosco, Brian Kelley, Ricardo Rodriguez, David Cosgrove, Riccardo Vianello, sriniker, Peter Gedeck, Gareth Jones, NadineSchneider, Eisuke Kawashima, Dan Nealschneider, Andrew Dalke, Matt Swain, Brian Cole, Samo Turk, Aleksandr Savelev, Alain Vaucher, Maciej Wójcikowski, Ichiru Take, Vincent~F. Scalfani, Rachel Walker, Daniel Probst, Kazuya Ujihara, Axel Pahl, guillaume godin, Juuso Lehtivarjo, tadhurst cdd, François Bérenger, and Jonathan Bisson. 2024.
\newblock \href {https://doi.org/10.5281/zenodo.13820100} {rdkit/rdkit: 2024\_09\_1 (q3 2024) release beta}.

\bibitem[{Li et~al.(2024)Li, Liu, Fan, Wei, Liu, Tang, and Li}]{Li2024molregpt}
Jiatong Li, Yunqing Liu, Wenqi Fan, Xiao-Yong Wei, Hui Liu, Jiliang Tang, and Qing Li. 2024.
\newblock \href {https://doi.org/10.1109/tkde.2024.3393356} {Empowering molecule discovery for molecule-caption translation with large language models: A chatgpt perspective}.
\newblock \emph{IEEE Transactions on Knowledge and Data Engineering}, page 1–13.

\bibitem[{Liu et~al.(2023)Liu, Zhang, Xia, Wu, Xie, Qin, Zhang, and Liu}]{liu-etal-2023-molxpt}
Zequn Liu, Wei Zhang, Yingce Xia, Lijun Wu, Shufang Xie, Tao Qin, Ming Zhang, and Tie-Yan Liu. 2023.
\newblock \href {https://doi.org/10.18653/v1/2023.acl-short.138} {{M}ol{XPT}: Wrapping molecules with text for generative pre-training}.
\newblock In \emph{Proceedings of the 61st Annual Meeting of the Association for Computational Linguistics (Volume 2: Short Papers)}, pages 1606--1616, Toronto, Canada. Association for Computational Linguistics.

\bibitem[{M.~Bran et~al.(2024)M.~Bran, Cox, Schilter, Baldassari, White, and Schwaller}]{bran2024augmenting}
Andres M.~Bran, Sam Cox, Oliver Schilter, Carlo Baldassari, Andrew~D. White, and Philippe Schwaller. 2024.
\newblock \href {https://doi.org/10.1038/s42256-024-00832-8} {Augmenting large language models with chemistry tools}.
\newblock \emph{Nature Machine Intelligence}, 6(5):525--535.

\bibitem[{Magister et~al.(2023)Magister, Mallinson, Adamek, Malmi, and Severyn}]{magister-etal-2023-teaching}
Lucie~Charlotte Magister, Jonathan Mallinson, Jakub Adamek, Eric Malmi, and Aliaksei Severyn. 2023.
\newblock \href {https://doi.org/10.18653/v1/2023.acl-short.151} {Teaching small language models to reason}.
\newblock In \emph{Proceedings of the 61st Annual Meeting of the Association for Computational Linguistics (Volume 2: Short Papers)}, pages 1773--1781, Toronto, Canada. Association for Computational Linguistics.

\bibitem[{Miller et~al.(2009)Miller, Vandome, and McBrewster}]{miller2009levenshtein}
Frederic~P Miller, Agnes~F Vandome, and John McBrewster. 2009.
\newblock Levenshtein distance: Information theory, computer science, string (computer science), string metric, damerau? levenshtein distance, spell checker, hamming distance.

\bibitem[{Nguyen et~al.(2024)Nguyen, Pham, Tran, and Manavalan}]{nguyen2024langmoldiff}
Nguyen Doan~Hieu Nguyen, Nhat~Truong Pham, Duong~Thanh Tran, and Balachandran Manavalan. 2024.
\newblock \href {https://openreview.net/forum?id=j9q7lurl7T} {Lang2mol-diff: A diffusion-based generative model for language-to-molecule translation leveraging {SELFIES} representation}.
\newblock In \emph{ACL 2024 Workshop Language + Molecules}.

\bibitem[{OpenAI and et~al.(2024)}]{openai2024gpt4technicalreport}
OpenAI and Josh~Achiam et~al. 2024.
\newblock \href {https://arxiv.org/abs/2303.08774} {Gpt-4 technical report}.
\newblock \emph{Preprint}, arXiv:2303.08774.

\bibitem[{Ouyang et~al.(2024)Ouyang, Zhang, Yan, Liu, Choi, Han, and Qin}]{ouyang2024structuredchemistryreasoninglarge}
Siru Ouyang, Zhuosheng Zhang, Bing Yan, Xuan Liu, Yejin Choi, Jiawei Han, and Lianhui Qin. 2024.
\newblock \href {https://openreview.net/forum?id=7R3pzxTSlg} {Structured chemistry reasoning with large language models}.
\newblock In \emph{Forty-first International Conference on Machine Learning, {ICML} 2024, Vienna, Austria, July 21-27, 2024}. OpenReview.net.

\bibitem[{Papineni et~al.(2002)Papineni, Roukos, Ward, and Zhu}]{papineni2002bleu}
Kishore Papineni, Salim Roukos, Todd Ward, and Wei-Jing Zhu. 2002.
\newblock Bleu: a method for automatic evaluation of machine translation.
\newblock In \emph{Proceedings of the 40th annual meeting of the Association for Computational Linguistics}, pages 311--318.

\bibitem[{Pei et~al.(2024)Pei, Wu, Gao, Zhu, and Yan}]{pei2024enhanced}
Qizhi Pei, Lijun Wu, Kaiyuan Gao, Jinhua Zhu, and Rui Yan. 2024.
\newblock \href {https://openreview.net/forum?id=Fib0IJt8YW} {Enhanced biot5+ for molecule-text translation: A three-stage approach with data distillation, diverse training, and voting ensemble}.
\newblock In \emph{ACL 2024 Workshop Language + Molecules}.

\bibitem[{Pei et~al.(2023)Pei, Zhang, Zhu, Wu, Gao, Wu, Xia, and Yan}]{pei-etal-2023-biot5}
Qizhi Pei, Wei Zhang, Jinhua Zhu, Kehan Wu, Kaiyuan Gao, Lijun Wu, Yingce Xia, and Rui Yan. 2023.
\newblock \href {https://doi.org/10.18653/v1/2023.emnlp-main.70} {{B}io{T}5: Enriching cross-modal integration in biology with chemical knowledge and natural language associations}.
\newblock In \emph{Proceedings of the 2023 Conference on Empirical Methods in Natural Language Processing}, pages 1102--1123, Singapore. Association for Computational Linguistics.

\bibitem[{Preuer et~al.(2018)Preuer, Renz, Unterthiner, Hochreiter, and Klambauer}]{preuer2018fcd}
Kristina Preuer, Philipp Renz, Thomas Unterthiner, Sepp Hochreiter, and G{\"u}nter Klambauer. 2018.
\newblock \href {https://doi.org/10.1021/acs.jcim.8b00234} {Fr{\'e}chet chemnet distance: A metric for generative models for molecules in drug discovery}.
\newblock \emph{Journal of Chemical Information and Modeling}, 58(9):1736--1741.

\bibitem[{Raffel et~al.(2020)Raffel, Shazeer, Roberts, Lee, Narang, Matena, Zhou, Li, and Liu}]{raffel2020t5}
Colin Raffel, Noam Shazeer, Adam Roberts, Katherine Lee, Sharan Narang, Michael Matena, Yanqi Zhou, Wei Li, and Peter~J. Liu. 2020.
\newblock \href {http://jmlr.org/papers/v21/20-074.html} {Exploring the limits of transfer learning with a unified text-to-text transformer}.
\newblock \emph{Journal of Machine Learning Research}, 21(140):1--67.

\bibitem[{Rogers and Hahn(2010)}]{rogers2010morgan}
David Rogers and Mathew Hahn. 2010.
\newblock Extended-connectivity fingerprints.
\newblock \emph{Journal of chemical information and modeling}, 50(5):742--754.

\bibitem[{Schneider et~al.(2015)Schneider, Sayle, and Landrum}]{schneider2015rdk}
Nadine Schneider, Roger~A Sayle, and Gregory~A Landrum. 2015.
\newblock Get your atoms in order - an open-source implementation of a novel and robust molecular canonicalization algorithm.
\newblock \emph{Journal of chemical information and modeling}, 55(10):2111--2120.

\bibitem[{Sun et~al.(2024)Sun, Luo, Gong, Lin, Shen, Guo, and Duan}]{sun2024itercot}
Jiashuo Sun, Yi~Luo, Yeyun Gong, Chen Lin, Yelong Shen, Jian Guo, and Nan Duan. 2024.
\newblock \href {https://doi.org/10.18653/v1/2024.findings-naacl.257} {Enhancing chain-of-thoughts prompting with iterative bootstrapping in large language models}.
\newblock In \emph{Findings of the Association for Computational Linguistics: NAACL 2024}, pages 4074--4101, Mexico City, Mexico. Association for Computational Linguistics.

\bibitem[{Touvron et~al.(2023)Touvron, Lavril, Izacard, Martinet, Lachaux, Lacroix, Rozière, Goyal, Hambro, Azhar, Rodriguez, Joulin, Grave, and Lample}]{touvron2023llamaopenefficientfoundation}
Hugo Touvron, Thibaut Lavril, Gautier Izacard, Xavier Martinet, Marie-Anne Lachaux, Timothée Lacroix, Baptiste Rozière, Naman Goyal, Eric Hambro, Faisal Azhar, Aurelien Rodriguez, Armand Joulin, Edouard Grave, and Guillaume Lample. 2023.
\newblock \href {https://arxiv.org/abs/2302.13971} {Llama: Open and efficient foundation language models}.
\newblock \emph{Preprint}, arXiv:2302.13971.

\bibitem[{Tran et~al.(2024)Tran, Pham, Nguyen, and Manavalan}]{tran2024mollangvlm}
Duong~Thanh Tran, Nhat~Truong Pham, Nguyen Doan~Hieu Nguyen, and Balachandran Manavalan. 2024.
\newblock \href {https://openreview.net/forum?id=ax8kYHfkNr} {Mol2lang-{VLM}: Vision- and text-guided generative pre-trained language models for advancing molecule captioning through multimodal fusion}.
\newblock In \emph{ACL 2024 Workshop Language + Molecules}.

\bibitem[{Wang et~al.(2023)Wang, Wei, Schuurmans, Le, Chi, Narang, Chowdhery, and Zhou}]{wang2023selfconsistency}
Xuezhi Wang, Jason Wei, Dale Schuurmans, Quoc~V Le, Ed~H. Chi, Sharan Narang, Aakanksha Chowdhery, and Denny Zhou. 2023.
\newblock \href {https://openreview.net/forum?id=1PL1NIMMrw} {Self-consistency improves chain of thought reasoning in language models}.
\newblock In \emph{The Eleventh International Conference on Learning Representations}.

\bibitem[{Wei et~al.(2022)Wei, Wang, Schuurmans, Bosma, ichter, Xia, Chi, Le, and Zhou}]{wei2023chainofthoughtpromptingelicitsreasoning}
Jason Wei, Xuezhi Wang, Dale Schuurmans, Maarten Bosma, brian ichter, Fei Xia, Ed~Chi, Quoc~V Le, and Denny Zhou. 2022.
\newblock Chain-of-thought prompting elicits reasoning in large language models.
\newblock In \emph{Advances in Neural Information Processing Systems}, volume~35, pages 24824--24837. Curran Associates, Inc.

\bibitem[{Weininger(1988)}]{weininger1988smiles}
David Weininger. 1988.
\newblock Smiles, a chemical language and information system. 1. introduction to methodology and encoding rules.
\newblock \emph{Journal of Chemical Information and Computer Sciences}, 28(1):31--36.

\bibitem[{White et~al.(2023)White, Hocky, Gandhi, Ansari, Cox, Wellawatte, Sasmal, Yang, Liu, Singh, and Peña~Ccoa}]{white2023assessment}
Andrew~D. White, Glen~M. Hocky, Heta~A. Gandhi, Mehrad Ansari, Sam Cox, Geemi~P. Wellawatte, Subarna Sasmal, Ziyue Yang, Kangxin Liu, Yuvraj Singh, and Willmor~J. Peña~Ccoa. 2023.
\newblock \href {https://doi.org/10.1039/D2DD00087C} {Assessment of chemistry knowledge in large language models that generate code}.
\newblock \emph{Digital Discovery}, 2:368--376.

\bibitem[{Xi et~al.(2023)Xi, Jin, Zhou, Zheng, Gao, Liu, Gui, Zhang, and Huang}]{xi-etal-2023-self}
Zhiheng Xi, Senjie Jin, Yuhao Zhou, Rui Zheng, Songyang Gao, Jia Liu, Tao Gui, Qi~Zhang, and Xuanjing Huang. 2023.
\newblock \href {https://doi.org/10.18653/v1/2023.findings-emnlp.762} {Self-{P}olish: Enhance reasoning in large language models via problem refinement}.
\newblock In \emph{Findings of the Association for Computational Linguistics: EMNLP 2023}, pages 11383--11406, Singapore. Association for Computational Linguistics.

\bibitem[{Zhang et~al.(2024)Zhang, Zhang, Li, hai zhao, Karypis, and Smola}]{zhang2024multimodal}
Zhuosheng Zhang, Aston Zhang, Mu~Li, hai zhao, George Karypis, and Alex Smola. 2024.
\newblock \href {https://openreview.net/forum?id=y1pPWFVfvR} {Multimodal chain-of-thought reasoning in language models}.
\newblock \emph{Transactions on Machine Learning Research}.

\end{thebibliography}

\clearpage
\newpage
\appendix
\appendix
\begin{center}{\bf {\LARGE Appendix}}\end{center}

\paragraph{Organization} The appendix is organized as follows: We first present the experimental details such as hyperparameters and prompts in \cref{appx: exp}. Then we provide the additional experimental results including the generated samples and additional ablation studies in \cref{appx: sample}. Next, we described the usage of AI assistants and scientific artifacts in \cref{appx: ai} and \cref{appx: artifact}, respectively.

\section{Experimental details}\label{appx: exp}

In this section, we provide the details of the experiments. All experimental code related to this paper is available at \url{https://github.com/yunhuijang/MSR} and our experiments are based on a single run. Additionally, we used the packages including rouge-score==0.1.2 and nltk==3.8.1.

\subsection{Structure information analysis}\label{appx: exp_anal}

Here, we describe the detailed settings for the analysis in \cref{subsec: failure}. To evaluate the understanding of two recent LLMs: Llama3-8B-Instruct \citep{touvron2023llamaopenefficientfoundation} and GPT-4o \citep{ openai2024gpt4technicalreport}, we prompt the LLMs to infer the structural information from the given molecular SMILES string and text description of the molecule.

\textbf{Prompts given text description of molecules.} First, we asked LLMs to infer the structural information from the text description of the molecule, with the prompt described in \cref{tab:appx_prompt_desc}.

\begin{figure}[h!]
  \centering
  \resizebox{\linewidth}{!}{
  \begin{tcolorbox}[
    colback=gray!5!white, 
    colframe=gray!60!black,
    title={
      \parbox[t]{\dimexpr\linewidth-4mm\relax}{%
        \ttfamily
        Prompts for M2S
      }
    }
  ]
\textbf{Head prompt: } You are now working as an excellent expert in chemistry and drug discovery.
    
    Given the SMILES representation of a molecule, your job is to predict the structural information of the molecule.
    
    The structural information of the molecule caption includes the molecular formula, the length of the longest carbon chain, the number of aromatic rings, the IUPAC name of all the rings, all the functional groups, the number of chiral centers with S and R configurations each, the molecular weight, the IUPAC name of the molecule.
    
    The functional group and ring IUPAC names should be on the list. The number of chiral centers should also be format \{"S": , "R": \}.

Your response should only be in the JSON format following \{"molecular formula": , "functional group": , "longest carbon chain length": , "aromatic ring": , "ring IUPAC name":, "chiral": \{"S": , "R": \}, "weight": , "IUPAC name": \}.

THERE SHOULD BE NO OTHER CONTENT INCLUDED IN YOUR RESPONSE. DO NOT CHANGE THE JSON KEY NAMES. 
  \\
  \\
  \textbf{Input prompt: } Input: \textcolor{blue}{[SMILES]} \\

  \end{tcolorbox}
  }
     \caption{\textbf{Prompts for structure information analysis given SMILES string.}}
    \label{tab:appx_prompt_desc}
\end{figure}

\textbf{Prompts given SMILES string.} Next, we asked LLMs to infer the structural information from the SMILES string, with the prompt described in \cref{tab:appx_prompt_smi}.

\begin{figure}[h!]
  \centering
  \resizebox{\linewidth}{!}{
  \begin{tcolorbox}[
    colback=gray!5!white, 
    colframe=gray!60!black,
    title={
      \parbox[t]{\dimexpr\linewidth-4mm\relax}{%
        \ttfamily
        Prompts for T2S
      }
    }
  ]
  \textbf{Head prompt: } You are now working as an excellent expert in chemistry and drug discovery.       
    
    Given the caption of a molecule, your job is to predict the structural information of the molecule.
    
    The molecule caption is a sentence that describes the molecule, which mainly describes the molecule\'s structures, properties, and production. 
    
    The structural information of the molecule caption includes the molecular formula, the length of the longest carbon chain, the number of aromatic rings, the IUPAC name of all the rings, all the functional groups, the number of chiral centers with S and R configurations each, the molecular weight, the IUPAC name of the molecule.
    
    The functional group and ring IUPAC names should be on the list. The number of chiral centers should also be format \{"S": , "R": \}.
    
    Your response should only be in the JSON format following \{"molecular formula": , "functional group": , "longest carbon chain length": , "aromatic ring": , "ring IUPAC name":, "chiral": \{"S": , "R": \}, "weight": , "IUPAC name": \}.

    THERE SHOULD BE NO OTHER CONTENT INCLUDED IN YOUR RESPONSE. DO NOT CHANGE THE JSON KEY NAMES. 
  \\
  \\
  \textbf{Input prompt: } Input: \textcolor{blue}{[Description]} \\

  \end{tcolorbox}
  }
    \caption{\textbf{Prompts for structure information analysis given text description.}}
    \label{tab:appx_prompt_smi}
\end{figure}

\FloatBarrier

\subsection{Molecule-to-text}\label{appx: exp_m2t}
Here, we describe the detailed settings for the experiments of molecule-to-text in \cref{subsec: mol_caption}. Note that we used four A100-80GB GPUs.

\begin{figure}[h!]
  \centering
  \resizebox{\linewidth}{!}{
  \begin{tcolorbox}[
    colback=gray!5!white, 
    colframe=gray!60!black,
    title={
      \parbox[t]{\dimexpr\linewidth-4mm\relax}{%
        \ttfamily
        Prompts for molecule2text
      }
    }
  ]
   \textbf{Head prompt: } You are now working as an excellent expert in chemistry and drug discovery.               
    
    Given the caption of a molecule, your job is to predict the SMILES representation of the molecule.
    
    The molecule caption is a sentence that describes the molecule, which mainly describes the molecule's structures, properties, and production.
    
    You can infer the molecule SMILES representation from the caption.   
    
    Before you infer the molecule SMILES representation, YOU SHOULD FIRST GENERATE the molecular formula, the length of the longest carbon chain, the number of aromatic rings, the IUPAC name of all the rings, all the functional groups, the number of chiral centers with S and R configurations each, the molecular weight, the IUPAC name of the molecule.
    \\
    \\
    
    Example 1:
    Instruction: Given the caption of a molecule, predict the SMILES representation of the molecule.
    
    Input: \textcolor{blue}{[Description][MSR]}
    
    Your output should be: \{"molecule": <SMILES>\}
    \\
    $\dots$
    \\
    Example $k$:
    Instruction: Given the caption of a molecule, predict the SMILES representation of the molecule.
    
    Input: \textcolor{blue}{[Description][MSR]}
    
    Your output should be: \{"molecule": <SMILES>\}
    \\
    \\
    You should FIRST generate the structural information following the examples above, and then provide the JSON format of the molecule SMILES based on that.
    
    NOTE THAT THE SMILES REPRESENTATION MUST BE IN THE JSON format above \{"molecule": \}. THERE SHOULD BE NO OTHER CONTENT INCLUDED IN YOUR JSON. DO NOT CHANGE THE JSON KEY NAME.
    \\
    \\
  \textbf{Input prompt: } Input: \textcolor{blue}{[Description]} \\

  \end{tcolorbox}
  }
    \caption{\textbf{Prompts for the generalist models in molecule captioning task.}}
    \label{tab:appx_prompt_m2t}
\end{figure}

\textbf{Hyperparameters.} The hyperparameters for the specialist models are provided in \cref{tab:appx_hyper_cap}. Note that MolT5-large was not trained for the same epochs as the other models due to limited resource.

\textbf{Prompts.} The prompts used for the generalist models are described in \cref{tab:appx_prompt_m2t}. We primarily followed the prompt presented by \cite{Li2024molregpt}.

\begin{table}[h]
    \centering
    \resizebox{\linewidth}{!}{
    \begin{tabular}{ccccc}
    \toprule
    Hyperparameter & MolT5-base & MolT5-large & ChemT5-small & ChemT5-base \\
    \midrule
    Batch size & 8 & 4 & 8 & 8 \\
    Learning rate & $2e^{-4}$ & $2e^{-4}$ & $6e^{-4}$ & $6e^{-4}$ \\
    Epochs & 250 & 220 & 250 & 250 \\
    Warmup ratio & 0 & 0 & 0.1 & 0.1 \\
    Weight decay & 0.01 & 0.01 & 0 & 0 \\    
    Lr scheduler & linear & linear & linear & linear  \\
    \bottomrule
    \end{tabular}}
    \caption{\textbf{Hyperparameters for molecule captioning.}}
    \label{tab:appx_hyper_cap}
\end{table}

\subsection{Text-to-molecule}\label{appx: exp_t2m}
Here, we described the detailed settings for the experiments of text-to-molecule in \cref{subsec: method_t2m}. Note that we also used four A100-80GB GPUs.

\textbf{Hyperparameters.} The hyperparameters for the reasoning and answering module for the specialist models are provided in \cref{tab:appx_hyper_gen_rea} and \cref{tab:appx_hyper_gen_ans}, respectively. Note that MolT5-large was not trained for the same number of epochs as the other models due to limited time constraints.

\begin{table}[h]
    \centering
     \resizebox{\linewidth}{!}{
    \begin{tabular}{ccccc}
    \toprule
    Hyperparameter & MolT5-base  & ChemT5-small & ChemT5-base \\
    \midrule
    Batch size & 8  & 8 & 8 \\
    Learning rate & $1e^{-3}$ & $6e^{-4}$ &  $6e^{-4}$ \\
    Epochs & 250  & 250 & 250 \\
    Warmup ratio & 0.1  & 0 & 0 \\
    Weight decay & 0  & 0 & 0 \\    
    Lr scheduler & cosine  & linear & linear \\
    \bottomrule
    \end{tabular}
    }
    \caption{\textbf{Hyperparameters for the reasoning module of text-based molecule generation.}}
    \label{tab:appx_hyper_gen_rea}
    \vspace{-0.1in}
\end{table}
\begin{table}[h!]
    \centering
     \resizebox{\linewidth}{!}{
    \begin{tabular}{ccccc}
    \toprule
    Hyperparameter & MolT5-base & MolT5-large & ChemT5-small & ChemT5-base \\
    \midrule
    Batch size & 8 & 4 & 8 & 8 \\
    Learning rate & $1e^{-3}$ & $1e^{-3}$ & $6e^{-4}$ &  $6e^{-4}$ \\
    Epochs & 250 & 140 & 250 & 250 \\
    Warmup ratio & 0.1 & 0.1 & 0 & 0 \\
    Weight decay & 0 & 0 & 0 & 0 \\    
    Lr scheduler & cosine & cosine & linear & linear \\
    \bottomrule
    \end{tabular}
    }
    \caption{\textbf{Hyperparameters for the answering module of text-based molecule generation.}}
    \label{tab:appx_hyper_gen_ans}
\end{table}

\textbf{Reasoning accuracy} The accuracies for molecular formula, longest carbon chain length, number of aromatic rings, chirality, and IUPAC names are computed by exact match. The accuracies for ring compounds and functional groups are computed by the ratio of intersection between the set of true and generated CoTs. Lastly, the accuracy for molecular weight is considered correct if the generated weight is within 95\% to 105\% of the true weight.

\textbf{Prompts.} The prompts used for the generalist models are described in \cref{tab:appx_prompt_t2m}. We also primarily followed the prompt presented by \cite{Li2024molregpt}.

\begin{figure}[h!]
  \centering
  \resizebox{\linewidth}{!}{
  \begin{tcolorbox}[
    colback=gray!5!white, 
    colframe=gray!60!black,
    title={
      \parbox[t]{\dimexpr\linewidth-4mm\relax}{%
        \ttfamily
        Prompts for text2molecule
      }
    }
  ]
   \textbf{Head prompt: } You are now working as an excellent expert in chemistry and drug discovery.   

Given the SMILES representation of a molecule and structural description of the molecule, your job is to predict the caption of the molecule.   

The molecule caption is a sentence that describes the molecule, which mainly describes the molecule's structures, properties, and production. \\
\\

Example 1: 

Instruction: Given the SMILES representation of a molecule, predict the caption of the molecule.

Input: \textcolor{blue}{[SMILES][MSR]}

Your output should be: \{"caption": <Description>\}

$\dots$

Example $k$: 

Instruction: Given the SMILES representation of a molecule, predict the caption of the molecule.

Input: \textcolor{blue}{[SMILES][MSR]}

Your output should be: \{"caption": <Description>\}\\
\\
Your response should only be in the JSON format above; THERE SHOULD BE NO OTHER CONTENT INCLUDED IN YOUR RESPONSE. \\
\\

  \textbf{Input prompt: } Input: \textcolor{blue}{[SMILES]}<\Algname> \\

  \end{tcolorbox}
  }
    \caption{\textbf{Prompts for generalist models in text-based molecule generation task.}}    \label{tab:appx_prompt_t2m}

\end{figure}

\FloatBarrier

\subsection{Ablation study}\label{appx:exp_ab}
Here, we describe the detailed settings for the ablation study. 

\textbf{Prompts for ChemCrow.} The prompts used for ChemCrow \citep{bran2024augmenting} are described in \cref{tab:appx_prompt_chem_m2t} and \cref{tab:appx_prompt_chem_t2m}. Notably, it was not able to apply few-shot learning for ChemCrow as it was not applicable as the original prompt proposed in ChemCrow does not include any few-shot setting. 

\begin{figure}[h]
  \centering
  \resizebox{\linewidth}{!}{
  \begin{tcolorbox}[
    colback=gray!5!white, 
    colframe=gray!60!black,
    title={
      \parbox[t]{\dimexpr\linewidth-4mm\relax}{%
        \ttfamily
        Prompts for molecule2text with ChemCrow
      }
    }
  ]
\textbf{Head prompt: } Given the SMILES representation of a molecule and structural description of the molecule, your job is to predict the caption of the molecule.

    "Final Answer" follows the format: Final Answer: \{"caption": \}
  \\
  \\
  \textbf{Input prompt: } The SMILES representation of the molecule is as follows: : \textcolor{blue}{[SMILES]} \\

  \end{tcolorbox}
  }
    \caption{\textbf{Prompts for molecule captioning with ChemCrow.}}\label{tab:appx_prompt_chem_m2t}
\end{figure}

\begin{figure}[h]
  \centering
  \resizebox{\linewidth}{!}{
  \begin{tcolorbox}[
    colback=gray!5!white, 
    colframe=gray!60!black,
    title={
      \parbox[t]{\dimexpr\linewidth-4mm\relax}{%
        \ttfamily
        Prompts for text2mol with ChemCrow
      }
    }
  ]
\textbf{Head prompt: } Given the caption of a molecule, your job is to predict the SMILES representation of the molecule.
    
    The molecule caption is a sentence that describes the molecule, which mainly describes the molecule's structures, properties, and production.

    You can infer the molecule SMILES representation from the caption. 
    
    "Final Answer" follows the format: Final Answer: \{"molecule": \}
  \\
  \\
  \textbf{Input prompt: } The caption is as follows: \textcolor{blue}{[Description]} \\

  \end{tcolorbox}
  }
    \caption{\textbf{Prompts for text-based molecule generation with ChemCrow.}}\label{tab:appx_prompt_chem_t2m}
\end{figure}

\newpage
\clearpage

\section{Additional experimental results}\label{appx: sample}
In this section, we provide additional experimental results including several concrete examples of generated samples.

\subsection{Molecule-to-text}\label{appx: samp_m2t}

Here, we show the samples of molecule captioning, i.e., generated text descriptions of given molecules in \cref{fig: appx_m2t}. Notably, we show the generated samples from base-sized models for fair comparison.

\begin{figure*}[h]
\vspace{0.2in}  
    \centering    \includegraphics[width=\linewidth]{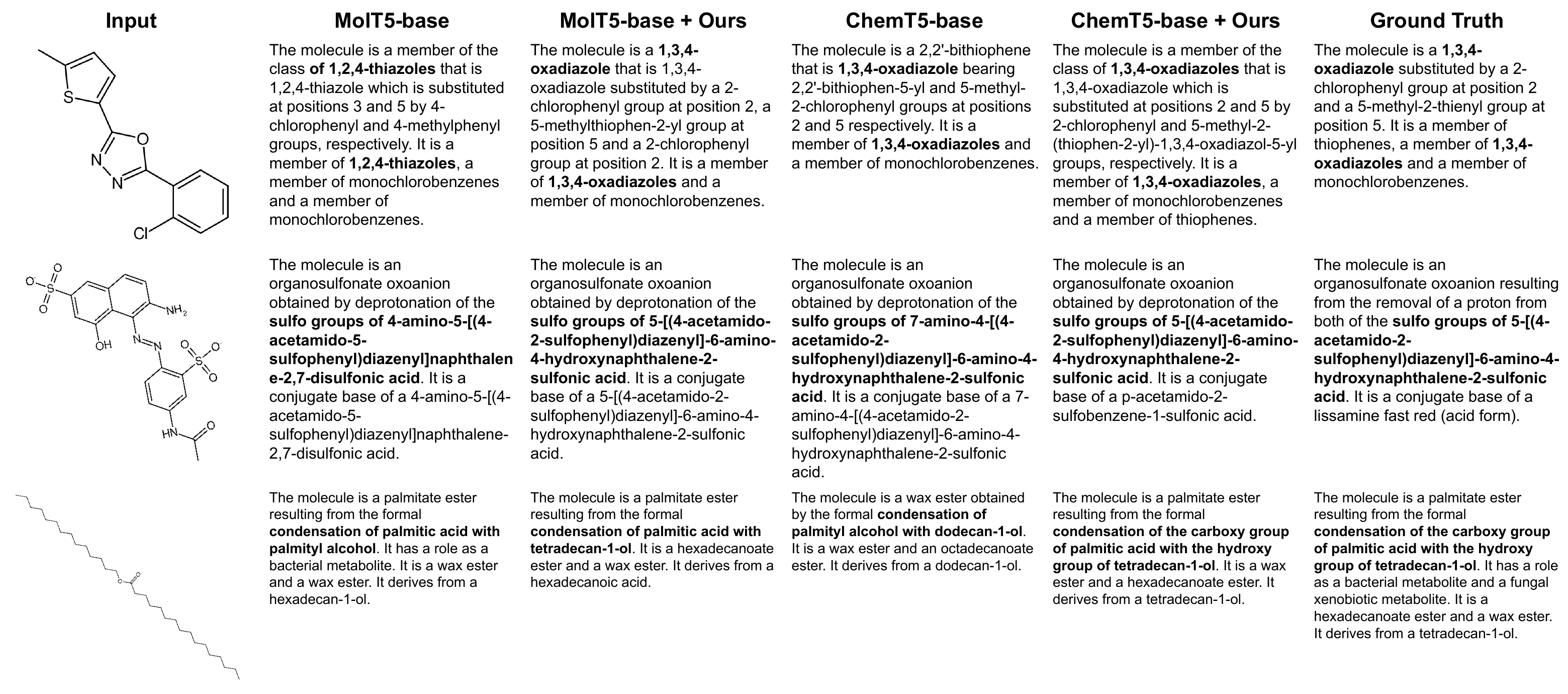}
    \caption{\textbf{The generated samples of molecule captioning.}}
    \label{fig: appx_m2t}
\end{figure*}

\subsection{Retrosynthesis}\label{appx: samp_retro}

Here, we show the samples of retrosynthesis, i.e., generated reactants of given product in \cref{fig: appx_retro}. 

\begin{figure*}[h]
\vspace{0.2in}  
    \centering    \includegraphics[width=\linewidth]{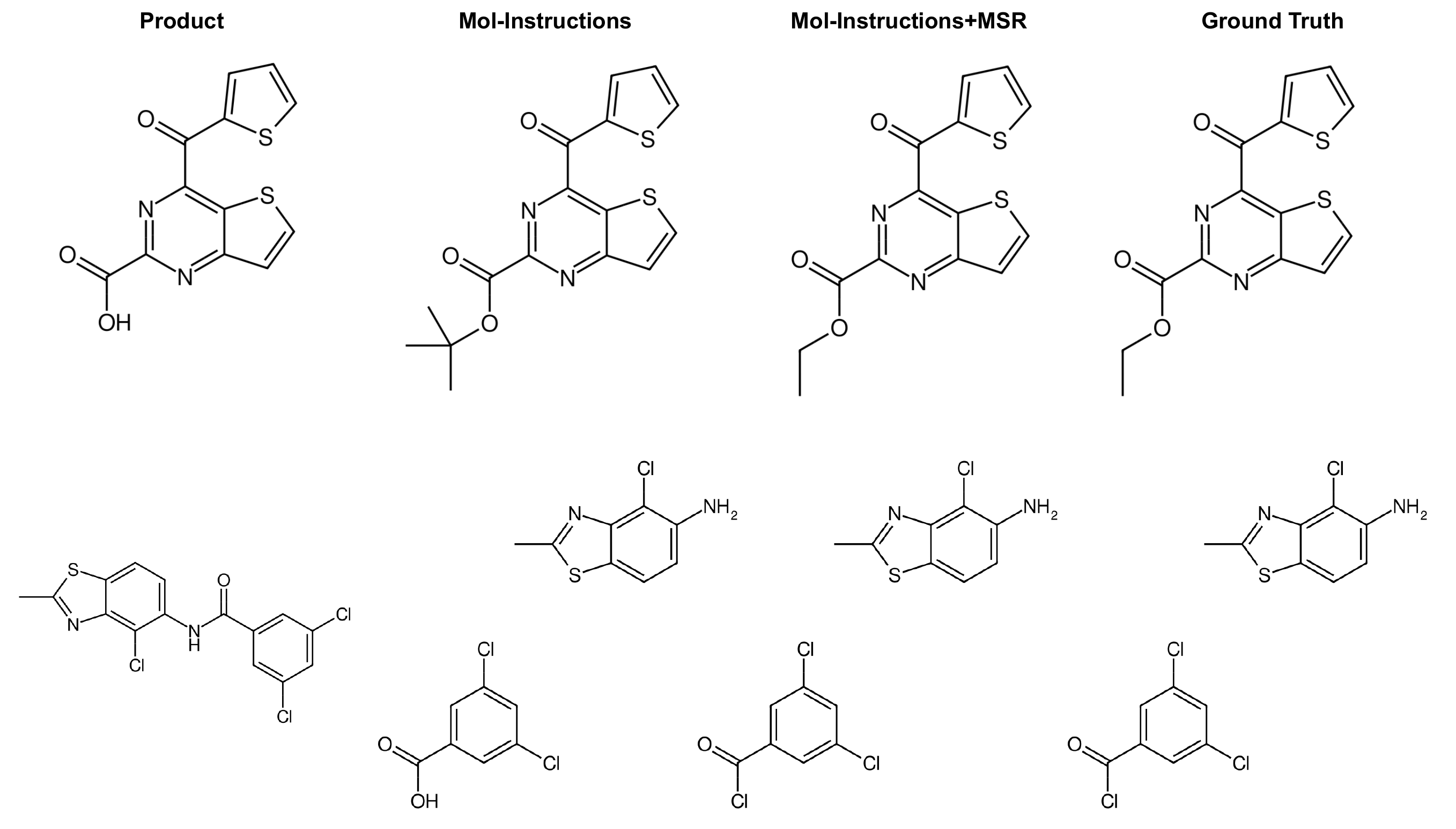}
    \caption{\textbf{The generated samples of retrosynthesis.}}
    \label{fig: appx_retro}
\end{figure*}

\subsection{Text-to-molecule}\label{appx: samp_t2m}

Here, we show the samples of text-based molecule generation, i.e., generated molecules for the given text description in \cref{fig: appx_t2m}. Notably, we show the generated samples from base-sized models for fair comparison.

\begin{figure*}[h]
    \centering
    \vspace{0.2in}    \includegraphics[width=\linewidth]{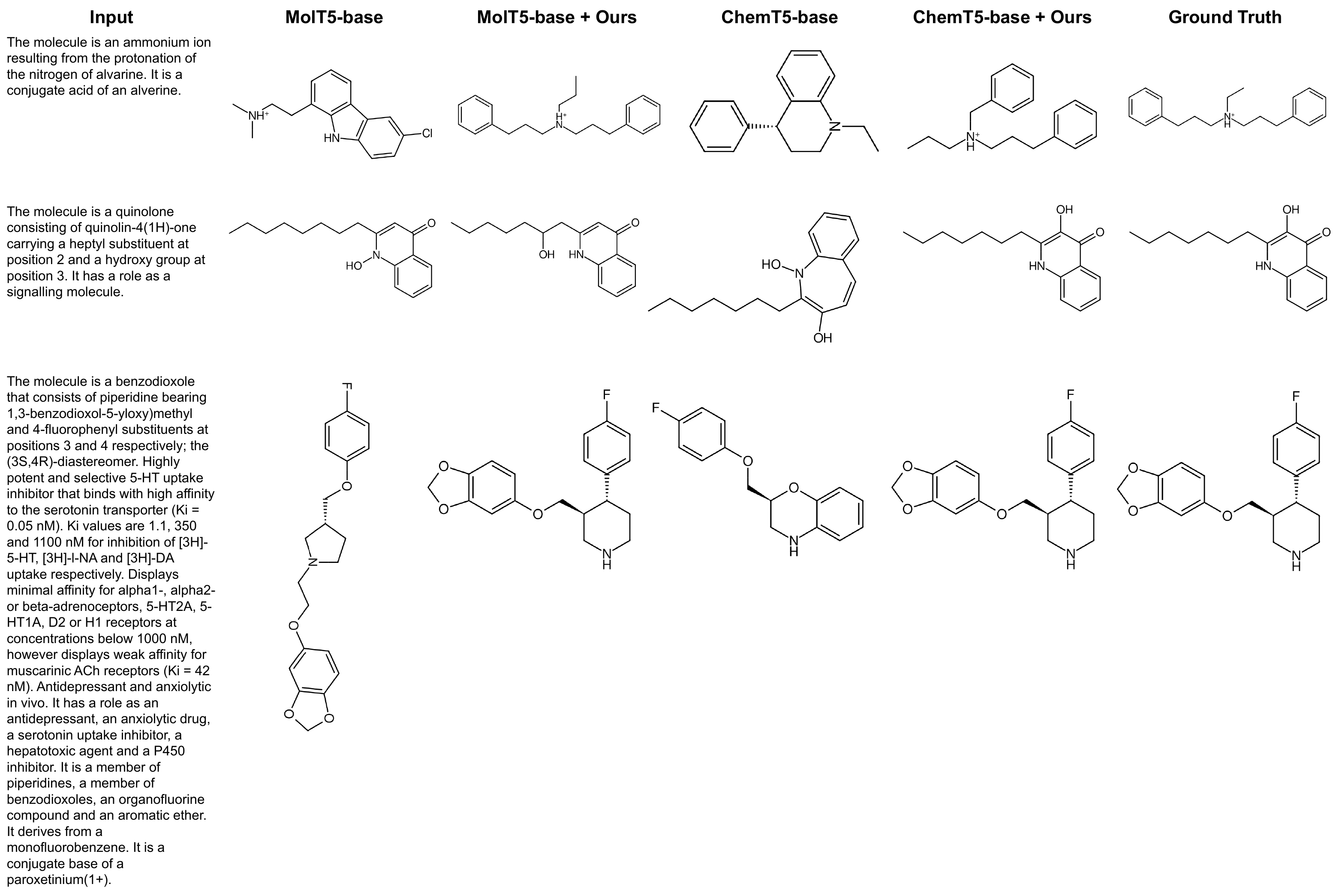}
    \caption{\textbf{The generated samples of text-based molecule generation.}}
    \label{fig: appx_t2m}
\end{figure*}

Additionally, we provide the results of generalist models in \cref{tab:appx_t2m_gen}. Note that it is natural to show no consistent enhancement for generalist models as they lack reasoning ability as shown in \cref{tab:reason}.

\definecolor{whitegray}{RGB}{243, 243, 243}

\begin{table}[h]
 
  \vspace{0.1in}
  \centering
  \resizebox{\linewidth}{!}{
\begin{tabular}{ccccccccc}
    \toprule[1.25pt]
Models & {\textbf{BL.} } & {\textbf{Ex.} } &{\textbf{Le.} $\downarrow$} &{\textbf{MA.}} &{\textbf{RDK}} &{\textbf{Mo.}} &{\textbf{FCD}$\downarrow$} &{\textbf{Val.}}\\
    \midrule[1.25pt]

    \rowcolor{whitegray} \multicolumn{9}{l}{\textit{Generalists (10-shot learning)}} \\
    \midrule
    Llama3  & 0.251  & 0.007  & 117.30  & 0.586  & 0.352  & 0.276  & 13.11  & 0.629  \\
    \quad +\Algname & 0.259  & 0.008  & 109.77  & 0.579  & 0.279  & 0.344  & 4.47  & 0.669  \\
    \midrule
    GPT-4o & 0.521  & 0.079  & 40.87  & 0.797  & 0.496  & 0.583  & 3.67  & 0.881  \\
    \quad +\Algname & 0.509  & 0.088  & 41.68  & 0.783 & 0.498  & 0.571  & 1.57  & 0.846\\
    \bottomrule[1.25pt]
    \end{tabular}
    }
     \caption{\textbf{Text-based Molecule Generation Performance for generalist models.}}\label{tab:appx_t2m_gen}
    \vspace{-0.25in}
\end{table}

\newpage

\subsection{Ablation study}\label{appx: samp_ab}

\textbf{Comparison to ChemCrow.}  To validate the efficacy of our \Algname, we compare our method with ChemCrow~\citep{bran2024augmenting}, which has employed CoTs for various chemical tasks. The comparative results are provided in \cref{tab: chemcrow_m2t} and \cref{tab: chemcrow_t2m}. One can observe that ChemCrow shows limited performance in both molecule captioning and text-based molecule generation tasks. \textcolor{black}{It is notable that the comparison may not be entirely appropriate, as ChemCrow is primarily designed for practical synthesis tasks, as the reviewer mentioned. Nevertheless, we included comparisons with ChemCrow to provide additional insights, as they share a similar motivation: enriching large language models (LLMs) with a chemistry-aware chain-of-thoughts.}

\setlength\intextsep{0pt}
\begin{table}[h]
\vspace{0.2in}

  \centering
  \resizebox{\linewidth}{!}{
    \begin{tabular}{ccccccc}
    \toprule
     Models & BL.-2 & BL.-4 & RO.-1 & RO.-2 & RO.-L & MET.  \\
    \midrule
     ChemCrow (GPT-4o) & 0.162 & 0.078 & 0.299 & 0.097 & 0.211 & 0.225  \\
     Ours (GPT-4o) & 0.249 & 0.139 & 0.386 & 0.179 & 0.300 & 0.303 \\
    Ours (ChemT5-base) & 0.639 & 0.560 & 0.687 & 0.553 & 0.626 & 0.657 \\
    \bottomrule
  \end{tabular}
  }
\caption{\textbf{Comparison to ChemCrow in molecule captioning.}} \label{tab: chemcrow_m2t}
\end{table}

\setlength\intextsep{0pt}
\begin{table}[h]
\vspace{0.2in}

  \centering
  \resizebox{\linewidth}{!}{
    \begin{tabular}{ccccccccc}
    \toprule
     Models & {\textbf{BL.} } & {\textbf{Ex.} } &{\textbf{Le.} $\downarrow$} &{\textbf{MA.}} &{\textbf{RDK}} &{\textbf{Mo.}} &{\textbf{FCD}$\downarrow$} &{\textbf{Val.}}\\
    \midrule
     ChemCrow (GPT-4o) & 0.306 & 0.194 & 56.46 & 0.772 & 0.632 & 0.555 & 2.31 & 0.851  \\
      Ours (GPT-4o) & 0.509 & 0.088 & 41.68 & 0.783 & 0.498 & 0.571 & 1.57 & 0.846 \\
    Ours (ChemT5-base) & 0.878 & 0.421 & 12.76 & 0.924 & 0.856 & 0.804 & 0.26 & 0.982 \\
   
    \bottomrule
  \end{tabular}
  }
\caption{\textbf{Comparison to ChemCrow in text-based molecule generation.}} \label{tab: chemcrow_t2m}
 \vspace{0.1in}
\end{table}

\newpage
\clearpage

\section{Usage of AI assistants}\label{appx: ai}
In preparing this work, we utilized AI-based writing assistants to refine sentence structure, correct grammatical errors, and enhance readability. These tools were employed only for rephrasing and language improvements, ensuring that the technical content, methodology, and experimental findings remained entirely authored by the researchers. The use of AI assistance was limited to editorial enhancements without influencing the originality or scientific contributions of the paper.

\newpage

\section{Scientific Artifacts}\label{appx: artifact}
\paragraph{The License for artifacts.} All datasets and software tools used in this work adhere to their respective licenses. Specifically, we employed publicly available datasets such as ChEBI-20 and L+M under their permitted usage terms. Additionally, external tools like RDKit were used following their open-source license. We release our trained models and code in \url{https://github.com/yunhuijang/MSR} under an appropriate open-source license to facilitate reproducibility.

\paragraph{Artifact use consistency with intended use.} The datasets and tools utilized in our study were used in accordance with their intended purpose. For example, ChEBI-20 and L+M datasets were originally developed for molecule captioning and generation tasks, aligning with our research objectives. Similarly, RDKit was employed for molecular structure analysis as intended by its developers.

\paragraph{Documentation of artifacts.} We provide details in \url{https://github.com/yunhuijang/MSR}.

\clearpage

\end{document}